\documentclass[10pt,twocolumn,letterpaper]{article}
\usepackage[utf8]{inputenc}
\usepackage{cvpr}
\usepackage{times}
\usepackage[sort,numbers]{natbib}
\usepackage[table,xcdraw]{xcolor}
\usepackage{amsmath}
\usepackage{amssymb}
\usepackage{bm}
\usepackage{booktabs}
\usepackage{graphicx}
\usepackage{enumitem}
\usepackage{mathtools}
\usepackage{microtype}
\usepackage{multirow}
\usepackage{tabu}
\usepackage{todonotes}
\usepackage{import}
\usepackage[pagebackref=true,breaklinks=true,letterpaper=true,colorlinks,bookmarks=false]{hyperref}
\usepackage{cleveref}

\setlist{noitemsep,leftmargin=*}%

\IfFileExists{/Users/vedaldi/.bash_profile}{
\hbadness=\maxdimen
\vbadness=\maxdimen
\vfuzz=30pt
\hfuzz=30pt
}{}

\cvprfinalcopy

\ifcvprfinal\fi
\makeatletter
\renewcommand{\paragraph}{%
  \@startsection{paragraph}{4}%
  {\z@}{0.5em}{-1em}%
  {\normalfont\normalsize\bfseries}%
}
\makeatother

\newcommand{\bx}{\bm{x}}
\newcommand{\bp}{\bm{p}}
\newcommand{\by}{\bm{y}}

\definecolor{tabgray}{HTML}{E9E8E8}

\presetkeys{todonotes}{inline,backgroundcolor=yellow}{}

\title{Self-supervised Learning of Interpretable Keypoints from Unlabelled Videos}
\author{%
Tomas Jakab\\
\footnotesize Visual Geometry Group\\[-0.35em]
\footnotesize University of Oxford\\[-0.35em]
\footnotesize {\tt tomj@robots.ox.ac.uk}
\and
Ankush Gupta\\
\footnotesize DeepMind, London\\[-0.35em]
\footnotesize {\tt ankushgupta@google.com}
\and
Hakan Bilen\\
\footnotesize School of Informatics\\[-0.35em]
\footnotesize University of Edinburgh\\[-0.35em]
\footnotesize {\tt hbilen@ed.ac.uk}
\and
Andrea Vedaldi\\
\footnotesize Visual Geometry Group\\[-0.35em]
\footnotesize University of Oxford\\[-0.35em]
\footnotesize {\tt vedaldi@robots.ox.ac.uk}}

\vbadness=\maxdimen

\makeatletter
\renewcommand{\paragraph}{%
  \@startsection{paragraph}{4}%
  {\z@}{0.5em}{-1em}%
  {\normalfont\normalsize\bfseries}%
}
\makeatother

\begin{document}
\maketitle
\thispagestyle{empty}
\begin{abstract}
We propose KeypointGAN, a new method for recognizing the pose of objects from a single image that 
for learning uses only unlabelled videos and a weak empirical prior on the object poses.
Video frames differ primarily in the pose of the objects they contain, 
so our method distils the pose information by analyzing the differences between frames.
The distillation uses a new dual representation of the geometry of objects as a set of 2D keypoints, and as a pictorial representation, \ie a skeleton image.
This has three benefits:
(1) it provides a tight `geometric bottleneck' which disentangles pose from appearance, 
(2) it can leverage powerful image-to-image translation networks to map between photometry and geometry,
and (3) it allows to incorporate empirical pose priors in the learning process.
The pose priors are obtained from unpaired data, such as from a different dataset or modality such as mocap, such that no annotated image is ever used in learning the pose recognition network.
In standard benchmarks for pose recognition for humans and faces, 
our method achieves state-of-the-art performance among methods that do not 
require any labelled images for training.
Project page: \url{http://www.robots.ox.ac.uk/~vgg/research/unsupervised_pose/}
\end{abstract}

\section{Introduction}\label{s:intro}

Learning with limited or no external supervision is one of the most significant open challenges in machine learning.
In this paper, we consider the problem of learning the 2D geometry of object categories such as humans and faces using raw videos and as little additional supervision as possible.
In particular, given as input a number of videos centred on the object, the goal is to learn automatically a neural network that can predict the \emph{pose} of the object from a single image.

\begin{figure}
\centering
\includegraphics[width=\linewidth]{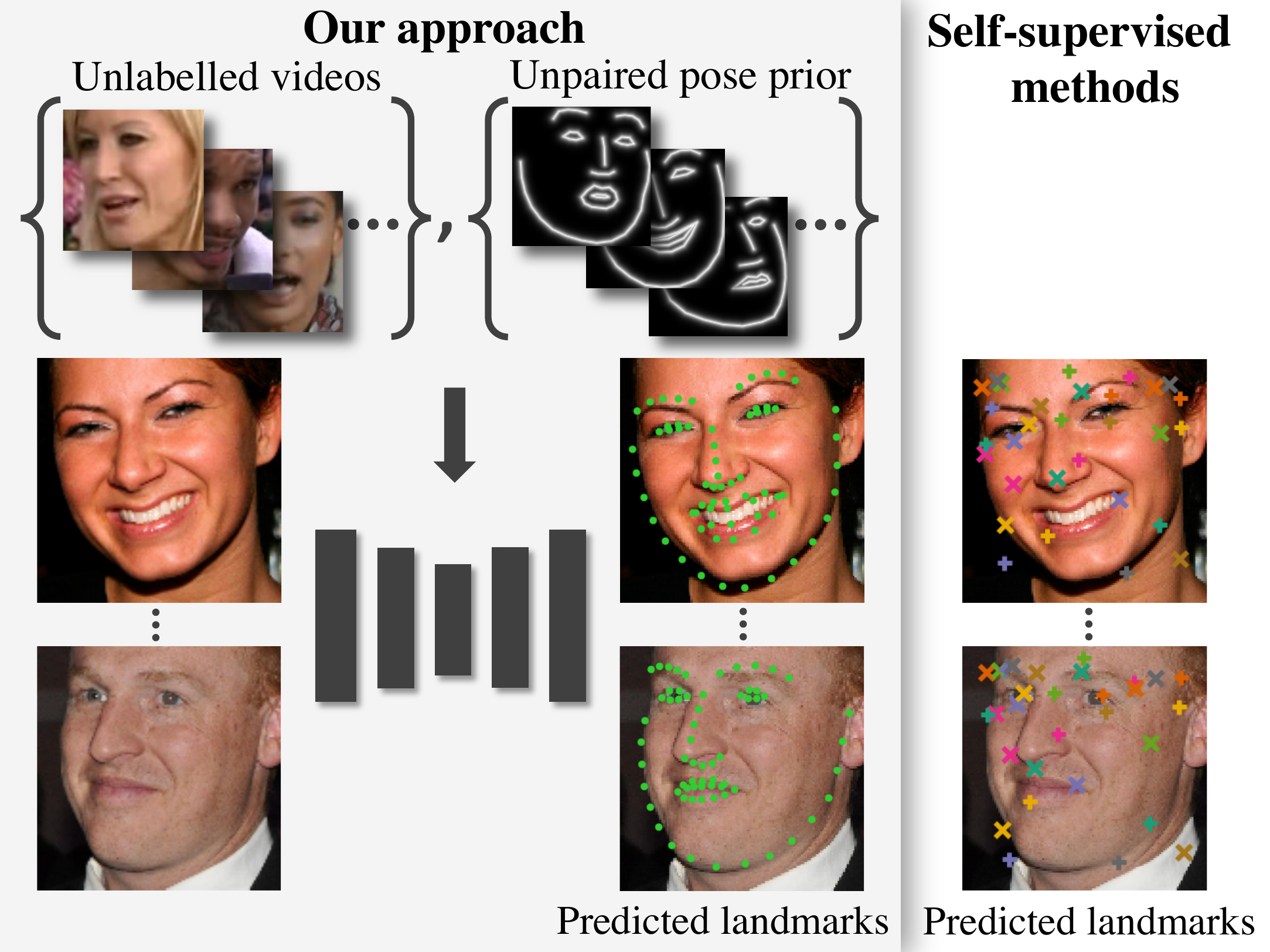}
\caption{\textbf{Learning landmark detectors from unpaired data.}
We learn to directly predict human-interpretable landmarks of an object using only unlabelled videos and a prior on the possible landmark configurations \textbf{[left]}.
The prior can be obtained from unpaired supervision or from a different modality, such as mocap data.
Our method, KeypointGAN, obtains state-of-the-art landmark detection performance for approaches that use unlabelled images for supervision.
In contrast, self-supervised landmark detectors~\cite{Thewlis17, zhang2018unsupervised, jakabunsupervised, lorenz2019unsupervised} can only learn to discover keypoints \textbf{[right]} that are not human-interpretable (predictions from~\cite{jakabunsupervised}) and require supervised post-processing.
}\label{f:splash}
\end{figure}

Learning from unlabelled images requires a suitable supervisory signal.
Recently, {}\cite{jakabunsupervised} noted that during a video an object usually maintains its intrinsic appearance but changes its pose.
Hence, the concept of pose can be learned by modelling the differences between video frames.
They formulate this as \emph{conditional image generation}.
They extract a small amount of information from a given target video frame via a tight bottleneck which retains pose information while discarding appearance.
For supervision, they reconstruct the target frame from the extracted pose, similar to an auto-encoder.
However, since pose alone does not contain sufficient information to reconstruct the appearance of the object, they also pass to the generator a second video frame from which the appearance can be observed.

In this paper, we also consider a conditional image generation approach, but we introduce a whole new design for the model and for the `pose bottleneck'.
In particular, we adopt a dual representation of pose as a set of 2D object coordinates, and as a pictorial representation of the 2D coordinates in the form of a skeleton image.
We also define a differentiable skeleton generator to map between the two representations.

This design is motivated by the fact that, by encoding pose labels as images we can leverage~\emph{powerful image-to-image translation networks}~\cite{zhu2017unpaired} to map between photometry and geometry.
In fact, the two sides of the translation process, namely the input image and its skeleton, are spatially aligned, which is well known to simplify learning by a Convolutional Neural Network (CNN)~\cite{zhu2017unpaired}.
At the same time, using 2D coordinates provides a very tight bottleneck that allows the model to efficiently separate pose from appearance.

The pose bottleneck is further controlled via a discriminator, learned adversarially.
This has the advantage of injecting prior information about the possible object poses in the learning process.
While acquiring this prior may require some supervision, this is separate from the unlabelled videos used to learn the pose recognizer --- that is, our method is able to leverage \emph{unpaired supervision}.
In this way, our method outputs poses that are directly interpretable.
We refer to our proposed method as \emph{\mbox{KeypointGAN}}.
By contrast, state-of-the-art self-supervised keypoint detectors~\cite{Thewlis17,jakabunsupervised,zhang2018unsupervised,Wiles18a,Shu2018} do not learn ``semantic'' keypoints and, in post-processing, they need at least some \emph{paired supervision} to output human-interpretable keypoints.
We highlight this difference in~\cref{f:splash}.

Overall, we make three significant contributions:
\begin{enumerate}
\item We introduce a new conditional generator design combining image translation, a new bottleneck using a dual representation of pose, and an adversarial loss which significantly improve recognition performance.
\item We learn, for the first time, to directly predict human-interpretable landmarks without requiring any labelled images.
\item We obtain state-of-the-art unsupervised landmark detection performance even when compared against methods that use paired supervision in post-processing.
\end{enumerate}

We test our approach using videos of people, faces, and cat images.
On standard benchmarks such as Human3.6M~\cite{h36m_pami} and 300-W~\cite{sagonas2016300}, 
we achieve state-of-the-art pose recognition performance for methods that learn only from unlabelled images.
We also probe generalization by testing whether the empirical pose prior can be extracted independently from the videos used to train the pose recognizer.
We demonstrate this in two challenging scenarios.
First, we use the mocap data from MPI-INF-3DHP~\cite{mehta2017monocular} as prior and we learn a human pose recognizer on videos from Human3.6M.
Second, we use the MultiPIE~\cite{sim2002cmu} dataset as prior to learn a face pose recognizer on VoxCeleb2~\cite{chung2018voxceleb2} videos, and achieve state-of-the-art facial keypoint detection performance on 300-W.

\begin{figure*}
\centering
\includegraphics[width=.9\textwidth]{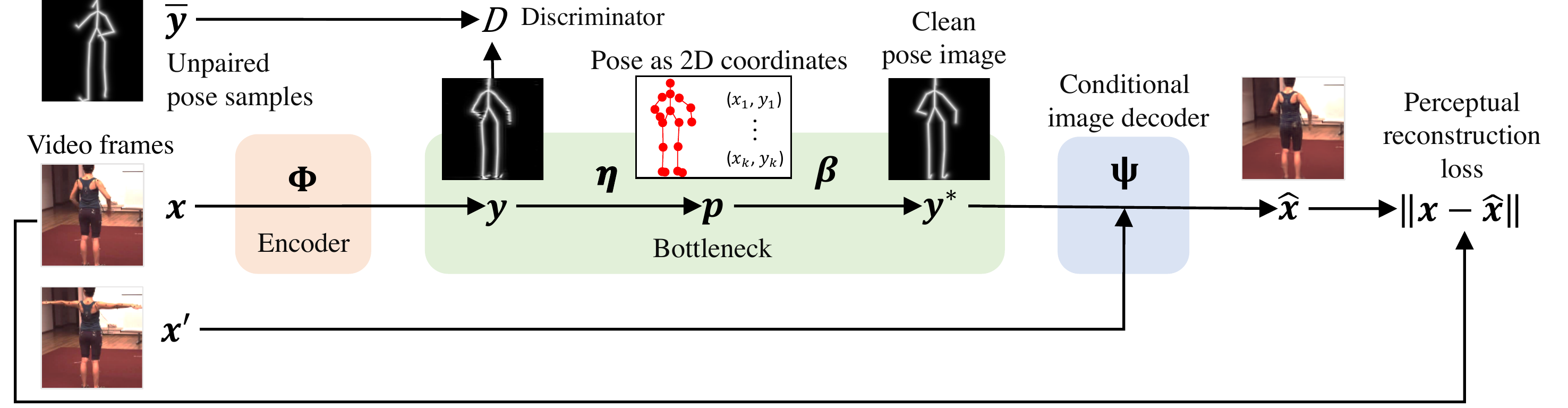}
\caption{\textbf{KeypointGAN architecture.}
We learn an encoder $\Phi$ that maps an image $\bx$ to its pose $\by$, represented as a skeleton image.
This is done via conditional auto-encoding, learning also a decoder $\Psi$ that reconstruct the input  $\bx$ from its pose $\by$ and a second auxiliary video frame $\bx'$.
A bottleneck $\beta\circ\eta$ is used to drop appearance information that may leak in the pose image $\by$.
A discriminator $D$ is used to match the distribution of predicted poses to a reference prior distribution, represented by unpaired pose samples $\bar\by$.
}
\label{f:schema}
\end{figure*}

\section{Related work}\label{s:relwork}

We consider pose recognition, intended as the problem of predicting the 2D pose of an object from a single image.
Approaches to this problem must be compared in relation to 
(1) the type of supervision, and (2) which priors they use.
There are three broad categories for supervision:
\emph{full supervision} when the training images are annotated with the same labels that one wishes to predict;
\emph{weak supervision} when the predicted labels are richer than the image annotations;
and \emph{no supervision} when there are no image annotations.
For the prior, methods can use a \emph{prior model} learned from any kind of data or supervision, 
an \emph{empirical prior}, or \emph{no prior} at all.

Based on this definition, our method is \emph{unsupervised} and uses an \emph{empirical prior}.
Next, we relate our work to others, dividing them by the type of supervision used (our method falls in the last category).

\paragraph{Full supervision.}

Several fully-supervised methods leverage large annotated datasets such as MS~COCO Keypoints~\cite{lin2014microsoft}, Human3.6M~\cite{h36m_pami}, MPII~\cite{andriluka20142d} and LSP~\cite{johnson2011learning}.
They generally do not use a separate prior as the annotations themselves capture one empirically.
Some methods use pictorial structures~\cite{felzenszwalb2005pictorial} to model the object poses~\cite{andriluka2009pictorial, sapp2010adaptive,yang2011articulated,pishchulin2013poselet, ouyang2014multi,ramakrishna2014pose}.
Others use a CNN to directly regress keypoint coordinates~\citep{toshev2014deeppose}, keypoint confidence maps~\citep{tompson2014joint}, or other relations between keypoints~\citep{chen2014articulated}.
Others again apply networks iteratively to refine heatmaps for single~\cite{wei2016convolutional,newell2016stacked,belagiannis2017recurrent, pfister2015flowing,carreira2016human,bulat2016human, tompson2015efficient} and multi-person settings~\cite{insafutdinov2016deepercut,cao2017realtime}.
Our method does not use any annotated image to learn the pose recognizer.

\paragraph{Weak supervision.}

A typical weakly-supervised method is the one of \citet{kanazawa2018end}:
they learn to predict dense 3D human meshes from sparse 2D keypoint annotations.
They use two priors: SMPL~\cite{loper2015smpl} parametric human mesh model,
and a prior on 3D poses acquired via adversarial learning from mocap data.
Analogous works include~\cite{tung2017adversarial,yang20183d,relativeposeBMVC18,geng193d-guided,gecer19ganfit,gerig18morphable,Sengupta18sfsnet,wang2019adversarial}.

All such methods use a prior trained on unpaired data, as we do.
However, they also use additional paired annotations such as 2D keypoints or relative depth relations~\citep{relativeposeBMVC18}.
Furthermore, in most cases they use a fully-fledged 3D prior such as SMPL human~\cite{loper2015smpl} or Basel face~\cite{bfm09} models, while we only use an empirical prior in the form of example 2D keypoints configurations.

\paragraph{No supervision.}

Other methods use no supervision, and some no data-driven prior either.
The works of~\cite{kanazawa2016warpnet,rocco2017convolutional,Wiles18a,Shu2018} learn to match pairs of images of an object, but they do not learn geometric invariants such as keypoints.
{}\cite{Thewlis17,Thewlis17a,thewlis2019unsupervised} do learn sparse and dense landmarks, also without any annotation.
The method of~\cite{Sundermeyer2018} does not use image annotations, but uses instead synthetic views of 3D models as prior, which we do not require.

Some of these methods use conditional image generation as we do.
Jakab \& Gupta et al.\cite{jakabunsupervised}, the most related, is described in the introduction.
\citet{zhang2018unsupervised,lorenz2019unsupervised} develop an auto-encoding formulation to discover landmarks as explicit structural representations for a given image and use them to reconstruct the original image.
\citet{Wiles18a,Shu2018} learn a dense deformation field for faces.
Our method differs from those in the particular nature of the model and geometric bottleneck;
furthermore, due to our use of a prior, we are able to learn out-of-the-box landmarks that are `semantically meaningful';
on the contrary, these approaches must rely on at least some paired supervision to translate between the unsupervised and `semantic' landmarks.
We also outperform these approaches in landmark detection quality.

\paragraph{Adversarial learning.}

Our method is also related to adversarial learning, which has proven to be useful in image labelling~\cite{ganin2015unsupervised,hoffman2017cycada,tzeng2015simultaneous,tzeng2017adversarial,gupta2018learning} and generation~\citep{goodfellow2014generative,zhu2017unpaired}, including bridging the domain shift between real and generated images.
Most relevant to our work, \citet{isola2017image} propose an image-to-image translation framework using paired data, while CycleGAN~\cite{zhu2017unpaired} can do so with unpaired data.
Our method also uses a image-to-image translation networks, but compared to CycleGAN our use of conditional image generation addresses the logical fallacy that an image-like label (a skeleton) does not contain sufficient information to generate a full image --- this issue is discussed in depth in~\cref{s:cyclegan}.

\paragraph{Appearance and geometry factorization.}

Recent methods for image generation conditioned on object attributes, like viewpoint~\cite{rhodin2018unsupervised}, pose~\cite{tran2017disentangled}, and hierarchical latents~\cite{singh2019finegan} have been proposed.
Our method allows for similar but more fine-grained conditional image generation, conditioned on an appearance image or object landmarks.
Many unsupervised methods for pose estimation~\cite{jakabunsupervised,zhang2018unsupervised,lorenz2019unsupervised,Wiles18a,Shu2018} share similar ability.
However, we can achieve more accurate and predictable image editing by manipulating semantic parts in the image through their corresponding landmarks.

\section{Method}\label{s:method}

Our goal is to learn a network $\Phi: \bx \mapsto \by$ that maps an image $\bx$ containing an object to its pose $\by$.
To avoid having to use image annotations, the network is trained using an auto-encoder formulation.
Namely, given the pose $\by = \Phi(\bx)$ extracted from the image, we train a decoder network $\Psi$ that reconstructs the image from the pose.
However, since pose lacks appearance information, this reconstruction task is ill posed.
Hence, we also provide the decoder with a \emph{different} image $\bx'$ of the same object to convey its appearance.
Formally, the image $\bx$ is reconstructed from the pose $\by$ and the auxiliary image $\bx'$ via a \emph{conditional decoder network}
\begin{equation}\label{e:autoencode}
  \bx = \Psi(\Phi(\bx),\bx').
\end{equation}
Unfortunately, without additional constraints, this formulation fails to learn pose properly~\cite{jakabunsupervised}.
The reason is that, given enough freedom, the encoder $\Phi(\bx)$ may simply decide to output a copy of the input image $\bx$, which allows it to trivially satisfy constraint~\eqref{e:autoencode} without learning anything useful (this issue is visualized in~\cref{s:cyclegan,f:cheat}).
The formulation needs a mechanism to force the encoder $\Phi$ to `distil' only pose information and discard appearance.

\begin{figure}[t]
\centering
\resizebox{.5\textwidth}{!}{
\begin{minipage}{\textwidth}
\LARGE
\begin{tabu} to 0.99\textwidth {X[c] X[c] X[c] X[c] X[c]}
  $\bx'$&  $\bx$&  $\hat\bx$& $\by$ & $\by^*$
\end{tabu}
\includegraphics[width=0.185\textwidth]{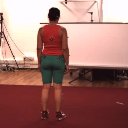}
\includegraphics[width=0.185\textwidth]{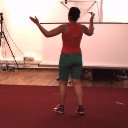}
\includegraphics[width=0.185\textwidth]{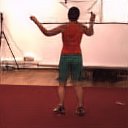}
\includegraphics[width=0.185\textwidth]{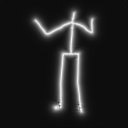}
\includegraphics[width=0.185\textwidth]{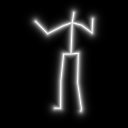}
\\
\includegraphics[width=0.185\textwidth]{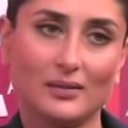}
\includegraphics[width=0.185\textwidth]{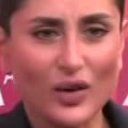}
\includegraphics[width=0.185\textwidth]{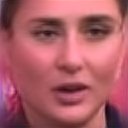}
\includegraphics[width=0.185\textwidth]{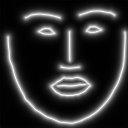}
\includegraphics[width=0.185\textwidth]{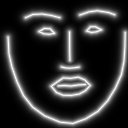}
\\
\begin{tabu} to 0.96\textwidth {X[c] X[c] X[c] X[c] X[c]}
style image & target image & reconstr. & skeleton image & clean skeleton.
\end{tabu}
\end{minipage}
}%
\caption{\textbf{Training data flow.} Data flowing through our model~(\cref{f:schema}) during training on the Human3.6M (human pose) and VoxCeleb2 (face) datasets. 
$\by,\by^*$ are our predictions.}
\label{f:overview}
\end{figure}

We make two key contributions to address these issues.
First, we introduce a \emph{dual representation of pose} as a vector of 2D keypoint coordinates and as a pictorial representation in the form of `skeleton' image~(\cref{s:dual}).
We show that this dual representation provides a tight bottleneck that distils pose information effectively while making it possible to implement the auto-encoder~\eqref{e:autoencode} using powerful image-to-image translation networks.

Our second contribution is to introduce an \emph{empirical prior} on the possible object poses (\cref{s:prior}).
In this manner, we can constrain not just the individual pose samples $\by$, but their \emph{distribution} $p(\by)$ as well.
In practice, the prior allows to use unpaired pose samples to improve accuracy and to learn an human-interpretable notion of pose that does not necessitate further learning to be used in applications.

\begin{figure}[]
\centering
{\large
$
\vcenter{\hbox{\includegraphics[width=0.18\linewidth]{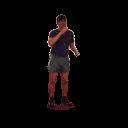}}}\overset{\Phi}{\rightarrow}
\vcenter{\hbox{\includegraphics[width=0.18\linewidth]{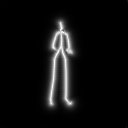}}}\overset{\Psi}{\rightarrow}
\vcenter{\hbox{\includegraphics[width=0.18\linewidth]{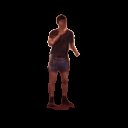}}}\hfill
\vcenter{\hbox{\includegraphics[width=0.18\linewidth]{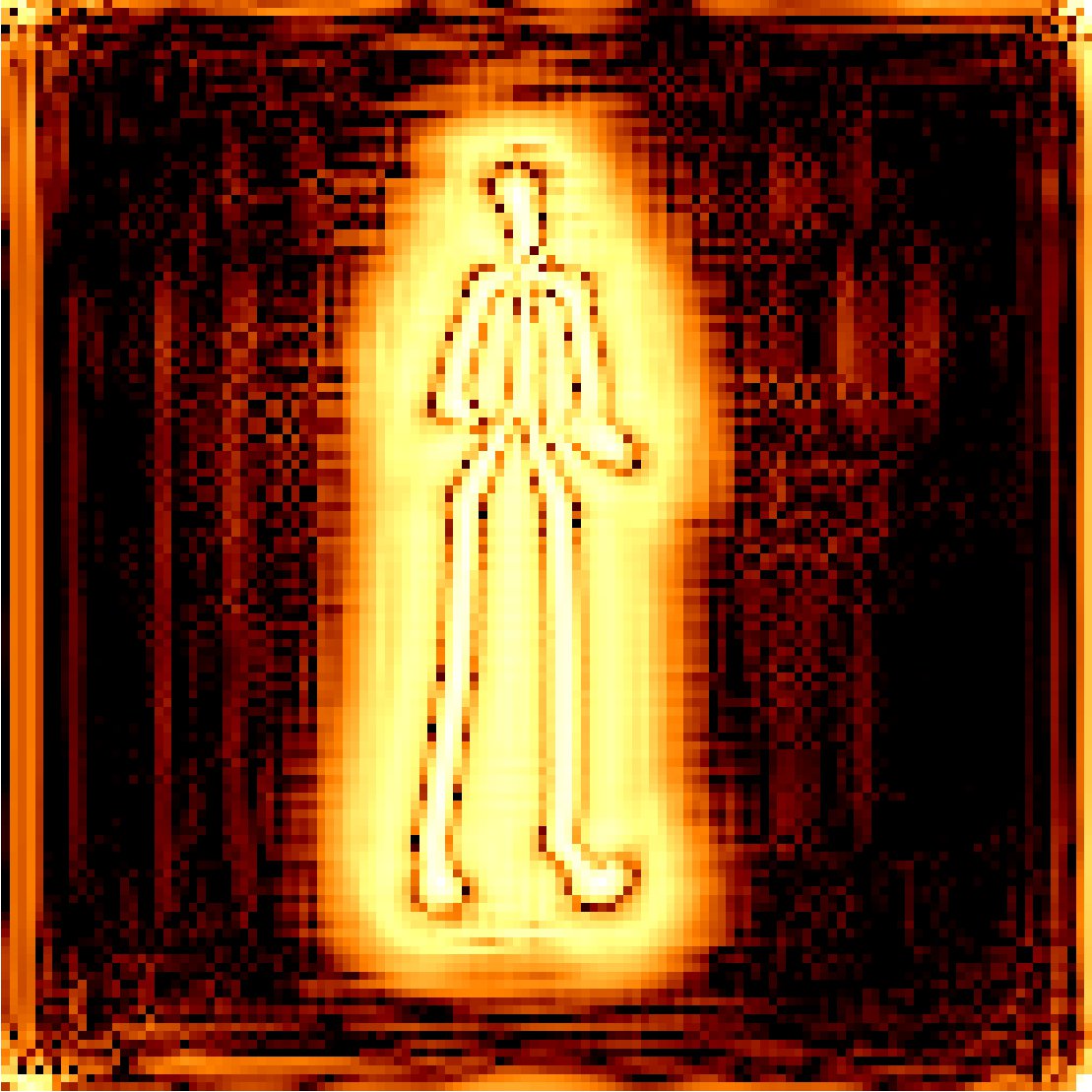}}}
$
}
\smallskip
\caption{\textbf{Leaking appearance in the pose representation.}
From left to right:
input image $\bx$, extracted skeleton image $\by = \Phi(\bx)$, and image reconstruction $\hat \bx = \Psi(\Phi(\bx))$.
In principle, it should not be possible to reconstruct the full image from only the skeleton, but the function $\Phi$ can `hide' the necessary information in a structured noise pattern, shown to the right as $\log \Phi(\bx)$.}\label{f:cheat}
\end{figure}

\subsection{Dual representation of pose \& bottleneck}\label{s:dual}
\label{s:bottleneck}

We consider a dual representation of the pose of an object as a vector of $K$ 2D \emph{keypoint coordinates} $\bp = (p_1,\dots,p_K)\in \Omega^K$ and as an \emph{image} $\by \in \mathbb{R}^{\Omega}$ containing a pictorial rendition of the pose as a skeleton (see~\cref{f:schema} for an illustration).
Here the symbol $\Omega= \{ 1,\dots,H \}\times \{ 1,\dots,W \}$ denotes a grid of pixel coordinates.

Representing pose as a set of 2D keypoints provides a tight bottleneck that preserves geometry but discards appearance information.
Representing pose as a skeleton image allows to implement the encoder and decoder networks as image translation networks.
In particular, the image of the object $\bx$ and of its skeleton $\by$ are \emph{spatially aligned}, which makes it easier for a CNN to map between them.

Next, we show how to switch between the two representations of pose.
We define the mapping $\by = \beta(\bp)$ from the coordinates $\bp$ to the skeleton image $\by$ \emph{analytically}.
Let $E$ be the set of keypoint pairs $(i,j)$ connected by a skeleton edge and let $u\in\Omega$ be an image pixel.
Then the skeleton image is given by:
\begin{equation}\label{e:skel}
   \beta(\bp)_u =
   \exp \left(
   - \gamma
   \min_{(i,j)\in E, r \in [0,1]} \|u - r\bp_i - (1-r) \bp_j\|^2
   \right)
\end{equation}
The differentiable function $\by = \beta(\bp)$ defines a distance field from line segments
that form the skeleton and applies an exponential fall off to generate an image. 
The visual effect is to produce a smooth line drawing of the skeleton.
We also train an inverse function $\bp = \eta(\by)$, implementing it as a neural network regressor (see supplementary for details).

Given the two maps $(\eta,\beta)$, we can use either representation of pose, as needed.
In particular, by using the pictorial representation $\by$, the encoder/pose recogniser can be written as an image-to-image translation network $\Phi: \bx \mapsto \by$ whose input $\bx \in \mathbb{R}^{3\times H\times W}$ and output $\by$ are both images.
The same is true for the conditional decoder $\Psi:(\by,\bx')\mapsto\bx$ of~\cref{e:autoencode}.

While image-to-image translation is desirable architecturally, the downside of encoding pose as an image $\by$ is that it gives the encoder $\Phi$ an opportunity to `cheat' and inject appearance information in the pose representation $\by$.
We can prevent cheating by exploiting the coordinate representation of pose to filter out any hidden appearance information form $\by$.
We do so by converting the pose image into keypoints and then back.
This amounts to substituting $\by = \beta \circ \eta (\by)$ in~\cref{e:autoencode}, which yields the modified auto-encoding constraint:
\begin{equation}\label{e:autoencode-bottleneck}
   \bx = \Psi(\beta \circ \eta \circ \Phi(\bx),\bx').
\end{equation}

\begin{figure*}
  \centering
  {\scriptsize
  \begin{tabu} to 0.99\linewidth {X[c] X[c] X[c]}
    Simplified Human3.6M~\cite{zhang2018unsupervised} &  Human3.6M &  PennAction
  \end{tabu}}
  \begin{minipage}[b]{57.5em}
    \includegraphics[width=0.092\linewidth]{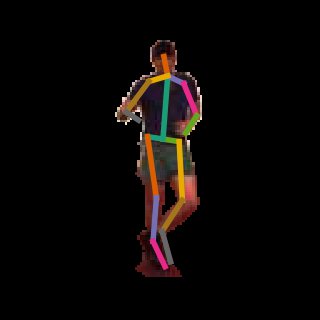}
    \includegraphics[width=0.092\linewidth]{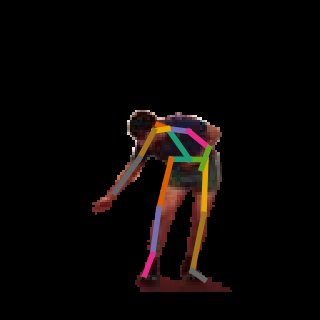}
    \includegraphics[width=0.092\linewidth]{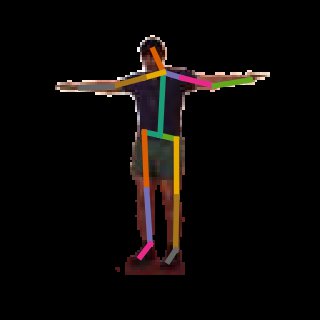}
    \includegraphics[width=0.092\linewidth]{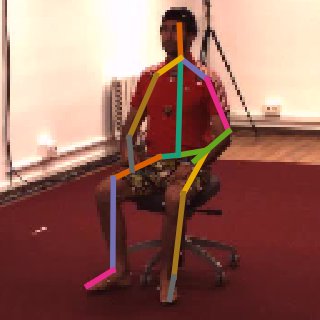}
    \includegraphics[width=0.092\linewidth]{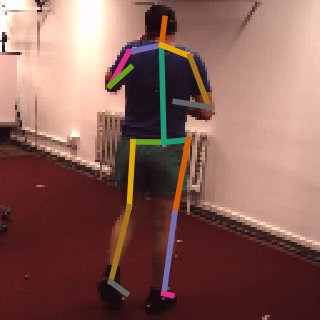}
    \includegraphics[width=0.092\linewidth]{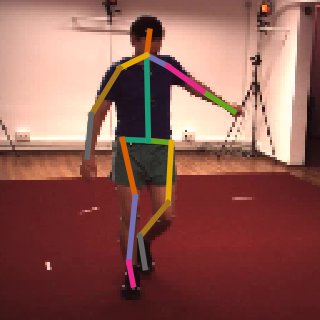}
    \includegraphics[width=0.092\linewidth]{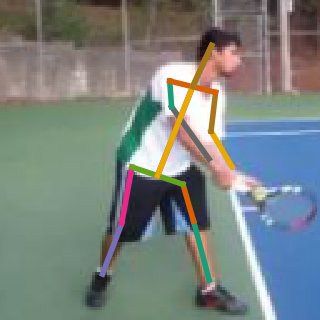}
    \includegraphics[width=0.092\linewidth]{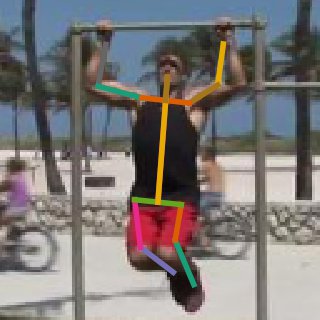}
    \includegraphics[width=0.092\linewidth]{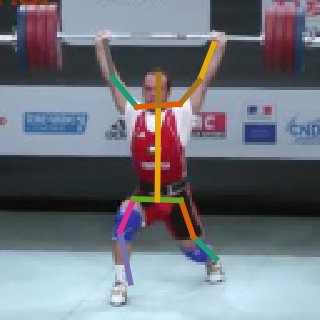}
  \end{minipage}
\caption{\textbf{Human pose predictions.}
2D keypoint predictions (visualised as connected limbs) on the simplified~\cite{zhang2018unsupervised} (with no background), full Human3.6M~\cite{h36m_pami}, and PennAction~\cite{zhang2013actemes} test sets.
Proposed KeypointGAN directly predicts human landmarks in complex poses without any additional supervision. 
More samples are included in the supplementary.}
\label{f:human-viz}
\end{figure*}

\subsection{Learning formulation \& pose prior}

\paragraph{Auto-encoding loss.}

In order to learn the auto-encoder~\eqref{e:autoencode-bottleneck}, we use a dataset of $N$ example pairs of video frames $\{(\bx_i,\bx_i')\}_{i=1}^{N}$.
Then the auto-encoding constraint~\eqref{e:autoencode-bottleneck} is enforced by optimizing a reconstruction loss.
Here we use a \emph{perceptual loss}:
\begin{equation}\label{e:perc}
  \mathcal{L}_\text{perc} =
  \frac{1}{N}
  \sum_{i=1}^N
   \| \Gamma(\hat \bx_i) - \Gamma(\bx_i)\|_2^2, 
\end{equation}
where $\hat \bx_i = \Psi(\beta\circ\eta\circ\Phi(\bx_i),\bx_i')$ is the reconstructed image, $\Gamma$ is a feature extractor.
Instead of comparing pixels directly, the perceptual loss compares features extracted from a standard network such as VGG~\cite{johnson2016perceptual,dosovitskiy2016generating,gatys2016image,bruna2015super}, and leads to more robust training.

\paragraph{Pose prior.}\label{s:prior}

In addition to the $N$ training image pairs $\{(\bx_i,\bx_i')\}_{i=1}^{N}$, we also assume to have access to $M$ sample poses $\{\bar \bp_j\}_{j=1}^{M}$.
Importantly, these sample poses are \emph{unpaired}, in the sense that they are not annotations of the training images.

We use the unpaired pose samples to encourage the predicted poses $\by$ to be plausible.
This is obtained by matching two distributions.
The reference distribution $q(\by)$ is given by the unpaired pose samples $\{\bar \by_j = \beta(\bar \bp_j)\}_{j=1}^{M}$.
The other distribution $p(\by)$ is given by the pose samples $\{\by_i = \Phi(\bx_i)\}_{i=1}^N$ predicted by the learned encoder network from the example video frames $\bx_i$.

The goal is to match $p(\by) \approx q(\by)$ in a distributional sense.
This can be done by learning a discriminator network $D(\by) \in [0, 1]$ whose purpose is to discriminate between the unpaired samples $\bar \by_j = \beta(\bar \bp_j)$ and the predicted samples $\by_i = \Phi(\bx_i)$.
Samples are compared by means of the \emph{difference adversarial loss} of~\cite{mao2017least}:
\begin{equation}\label{e:adversarial}
\mathcal{L}_\text{disc}(D) =
\frac{1}{M}
\sum_{j=1}^M 
D(\bar\by_j)^2 
+
\frac{1}{N}
\sum_{i=1}^N
(1-D(\by_i))^2.
\end{equation}

In addition to capturing plausible poses, the pose discriminator $D(\by)$ \emph{also} encourages the images $\by$ to be `skeleton-like'.
The effect is thus similar to the bottleneck introduced in~\cref{s:bottleneck} and one may wonder if the discriminator makes the bottleneck redundant.
The answer, as shown in~\cref{s:cyclegan,s:exp}, is negative: both are needed.

\paragraph{Overall learning formulation.}

Combining losses~\eqref{e:perc} and~\eqref{e:adversarial} yields the overall objective:
\begin{equation}\label{e:loss-overall}
   \mathcal{L}(\Phi,\Psi,D)
   =
   \lambda \mathcal{L}_\text{disc}(D, \Phi)
   +
   \mathcal{L}_\text{perc}(\Psi, \Phi),
\end{equation}
where $\lambda$ is a loss-balancing factor.
The components of this model, \emph{KeypointGAN}, and their relations are illustrated in~\cref{f:schema}.
Similar to any adversarial formulation, \cref{e:loss-overall} is minimized w.r.t.~$\Phi,\Psi$ and maximised w.r.t.~$D$.

\paragraph{Details.}

The functions $\Phi$, $\Psi$, $\eta$ and $D$ are implemented as convolutional neural networks.
The auto-encoder functions $\Phi$  and $\Psi$ and the discriminator $D$ are trained by optimizing the objective in~\cref{e:loss-overall} ($\eta$ is pre-trained using unpaired landmarks, for details see supplementary).
Batches are formed by sampling random pairs of \emph{video frames} $(\bx_i,\bx_i')$ and unpaired pose $\bar \by_j$ samples.
When sampling from \emph{image datasets} (instead of videos), we generate image pairs as $(g_1(\bx_i),g_2(\bx_i))$ by applying random thin-plate-splines $g_1,g_2$ to training samples $\bx_i$.
All the networks are trained from scratch.
Architectures and training details are in the supplementary.

\section{Relation to image-to-image translation}\label{s:cyclegan}

Our method is related to unpaired image-to-image translation, of which CycleGAN~\cite{zhu2017unpaired} is perhaps the best example, but with two key differences:
(a) it has a bottleneck (\cref{s:bottleneck}) that prevents leaking appearance information into the pose representation $\by$, and
(b) it reconstructs the image $\bx$ conditioned on a second image $\bx'$.
We show in the experiments that these changes are critical for pose recognition performance, and conduct a further analysis here.

First, consider what happens if we drop both changes (a) and (b), thus making our formulation more similar to CycleGAN\@.
In this case,~\cref{e:autoencode} reduces to $\bx = \Psi(\Phi(\bx))$.
The trivial solution of setting  both $\Phi$ and $\Psi$ to the identity functions is only avoided due to the discriminator loss~\eqref{e:adversarial}, which encourages $\by = \Phi(\bx)$ to look like a skeleton (rather than a copy of $\bx$).
In theory, then, this problem should be ill-posed as the pose $\by$ should not have sufficient information to recover the input image $\bx$.
However, the reconstructions from such a network still look reasonably good (see \cref{f:cheat}).
A closer look at logarithm of the generated skeleton $\by$, reveals that CycleGAN `cheats' by leaking appearance information via subtle patterns in $\by$.
By contrast, our bottleneck significantly limits leaking appearance in the pose image and thus its ability to reconstruct $\bx = \Psi(\beta\circ\eta\circ\Phi(\bx))$ from a single image;
instead, reconstruction is achieved by injecting the missing appearance information via the auxiliary image $\bx'$ using a conditional image decoder~(\cref{e:autoencode-bottleneck}).

\section{Experiments}\label{s:exp}

We evaluate our method, \emph{KeypointGAN}, on the task of 2D landmark detection for human pose~(\cref{s:x-human}), faces~(\cref{s:faces}), and cat heads~(\cref{s:animals}) and outperform state-of-the-art methods~(\cref{tab:face-sota,tab:simple-pose-sota,tab:pose-sota}) on these tasks.
We examine the relative contributions of components of our model in an ablation study~(\cref{s:x-abal}).
We study the effect of reducing the number of pose samples used in the empirical prior~(\cref{s:x-sample}).
Finally, we demonstrate image generation and manipulation conditioned on appearance and pose~(\cref{s:factor}).

\paragraph{Evaluation.}

KeypointGAN directly outputs predictions for keypoints that are human-interpretable.
In contrast, self-supervised methods~\cite{Thewlis17a,Thewlis17,Wiles18a,thewlis2019unsupervised,zhang2018unsupervised,lorenz2019unsupervised,jakabunsupervised} predict only \emph{machine-interpretable} keypoints, as illustrated in~\cref{f:splash}, and require at least some example images with paired keypoint annotations in order to learn to convert these landmarks to human-interpretable ones for benchmarking or for applications.
We call this step \emph{supervised post-processing}.
KeypointGAN does not require this step, but we \emph{also} include this result for a direct comparison with previous methods.

\subsection{Human pose}\label{s:x-human}

\paragraph{Datasets.}
\emph{Simplified Human3.6M} introduced by \citet{zhang2018unsupervised} for evaluating unsupervised pose recognition, contains 6 activities in which human bodies are mostly upright; 
it comprises 800k training and 90k testing images.
\emph{Human3.6M}~\cite{h36m_pami} is a large-scale dataset that contains 3.6M accurate 2D and 3D human pose annotations for 17 different activities, imaged under 4 viewpoints and a static background.
For training, we use subjects 1, 5, 6, 7, and 8, and subjects 9 and 11 for evaluation, as in~\cite{villegas2017learning}.
\emph{PennAction}~\cite{zhang2013actemes} contains 2k challenging consumer videos of 15 sports categories.
\emph{MPI-INF-3DHP}~\cite{mehta2017monocular} is a mocap dataset containing 8 subjects performing 8 activities in complex exercise poses. There are 28 joints annotated.

We split datasets into two \emph{disjoint} parts for sampling image pairs $(\bx,\bx')$ (cropped to the provided bounding boxes), and skeleton prior respectively to ensure that the pose data does not contain labels corresponding to the training images.
For the Human3.6M datasets we split the videos in half, while for PennAction we split in half the set of videos from each action category.
We also evaluate the case when images and skeletons are sampled from different datasets and for this purpose we use the MPI-INF-3DHP mocap data.

\paragraph{Evaluation.}

We report 2D landmark detection performance on the simplified and original Human3.6M datasets.
For Simplified Human3.6M, we follow the standard protocol of~\cite{zhang2018unsupervised} and report the error for all 32 joints normalized by the image size.
For Human3.6M, we instead report the mean error in pixels over 17 of the 32 joints~\cite{h36m_pami}.
To demonstrate learning from unpaired prior, we consider two settings for sourcing the images and the prior.
In the first setting, we use \emph{different} datasets for the two, and sample images from Human3.6M and poses from MPI-INF-3DHP\@.
In the second setting, we use instead two \emph{disjoint} parts of the \emph{same} dataset Human3.6M for both images and poses.
When using MPI-INF-3DHP dataset as the prior, we predict 28 joints, but use 17 joints that are common with Human3.6M for evaluation.
We train KeypointGAN from scratch and compare its performance with both supervised and unsupervised methods.

\paragraph{Results.}
\Cref{tab:simple-pose-sota} reports the results on Simplified Human3.6M.
As in previous self-supervised works~\cite{zhang2018unsupervised,Thewlis17}, we compare against the supervised baseline by Newell~\etal~\cite{newell2016stacked}.
Our model outperforms all the baselines~\cite{lorenz2019unsupervised,zhang2018unsupervised,Thewlis17} \emph{without} the supervised post-processing used by the others.
\begin{table}[t]
  \setlength{\tabcolsep}{2pt}
  \centering
  \resizebox{\linewidth}{!}{
    \begin{tabular}{llcccccc}
      \toprule
      \multicolumn{1}{l|}{Method}                                   & \multicolumn{1}{l|}{all}           & wait          & pose          & greet         & direct        & discuss       & walk          \\ \hline
      \multicolumn{8}{c}{\textbf{\emph{fully supervised}}}                                                                                                                                                                     \\
      \multicolumn{1}{l|}{Newell~\etal~\cite{newell2016stacked}}              & \multicolumn{1}{l|}{\textbf{2.16}} & \textbf{1.88} & \textbf{1.92} & \textbf{2.15} & \textbf{1.62} & \textbf{1.88} & \textbf{2.21} \\ \hline
      \multicolumn{8}{c}{\textbf{\emph{self-supervised + supervised post-processing}}}                                                                                                                                   \\
      \multicolumn{1}{l|}{Thewlis~\etal~\cite{Thewlis17}}           & \multicolumn{1}{l|}{7.51}          & 7.54          & 8.56          & 7.26          & 6.47          & 7.93          & 5.40          \\
      \multicolumn{1}{l|}{Zhang~\etal~\cite{zhang2018unsupervised}} & \multicolumn{1}{l|}{4.14}          & 5.01          & 4.61          & 4.76          & 4.45          & 4.91          & 4.61          \\
      \multicolumn{1}{l|}{Lorenz~\etal~\cite{lorenz2019unsupervised}} & \multicolumn{1}{l|}{2.79}        & ---            & ---            & ---            & ---            & ---            & ---          \\
      \hline
      \multicolumn{8}{c}{\textbf{\emph{self-supervised (no post-processing)}}}
      \\
      \multicolumn{1}{l|}{KeypointGAN (ours)}       & \multicolumn{1}{l|}{\textbf{2.73}} & \textbf{2.66} & \textbf{2.27} & \textbf{2.73} & \textbf{2.35} & \textbf{2.35} & \textbf{4.00} \\ 
      \bottomrule
      \end{tabular}}

      \smallskip																		   
\caption{{\bf Human landmark detection (Simplified H3.6M).} Comparison with state-of-the-art methods for human landmark detection on the Simplified Human3.6M dataset~\cite{zhang2018unsupervised}.
We report $\%$-MSE normalised by image size for each activity.}\label{tab:simple-pose-sota}
\end{table}

\Cref{tab:pose-sota} summarises our results on the original Human3.6M test set.
Here we also compare against the supervised baseline~\cite{newell2016stacked} and the self-supervised method of~\cite{jakabunsupervised}.
Our model outperforms the baselines in this test too.

It may be surprising that KeypointGAN outperforms the supervised baseline.
A possible reason is the limited number of supervised examples, 
which causes the supervised baseline to overfit.
This can be noted by comparing the training / test errors: 14.61 / 19.52 for supervised hourglass and  13.79 / 14.46 for our method.

\begin{figure}[h]
\includegraphics[height=0.192\linewidth,width=0.192\linewidth]{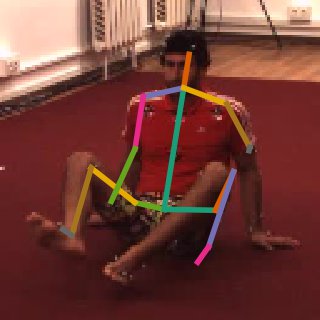}
\includegraphics[height=0.192\linewidth,width=0.192\linewidth]{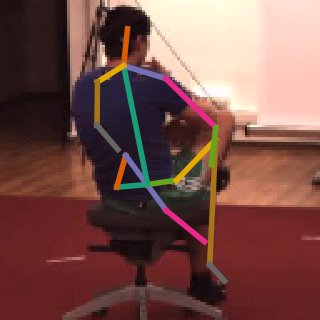}
\includegraphics[height=0.192\linewidth,width=0.192\linewidth]{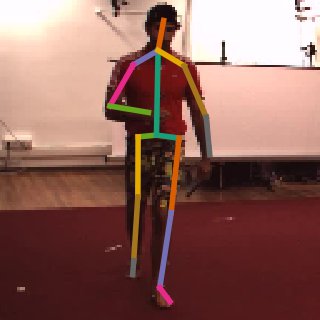}
\includegraphics[height=0.192\linewidth,width=0.192\linewidth]{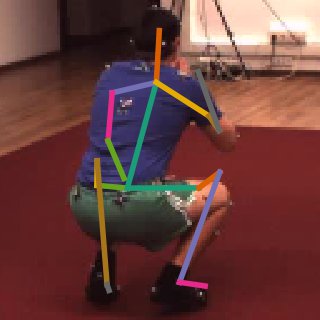}
\includegraphics[height=0.192\linewidth,width=0.192\linewidth]{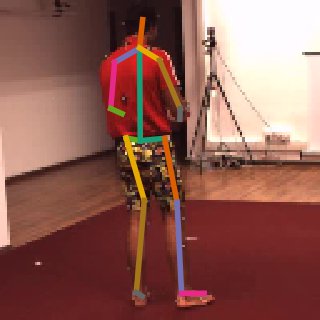}
\caption{\textbf{Limitations.}
\textbf{[1-2]}~complex human poses like sitting are challenging to learn
from a weak pose prior,
\textbf{[3]}~it could be difficult to disambiguate the sides due to bilateral symmetry,
\textbf{[4-5]}~occlusions are difficult to handle.
}\label{f:human-lim}
\end{figure}

When poses are sampled from a different dataset  (MPI-INF-3DHP) than the images (Human3.6M), the error is higher at 18.94 (but still better than the supervised alternative).
This increase is due to the domain gap between the two datasets.
\Cref{f:human-viz} shows some qualitative examples.
Limitations of KeypointGAN are highlighted in~\cref{f:human-lim}.

\begin{table}[t]
  \centering
  \begin{tabular}{@{}lr@{}}
    \toprule
    Method                                                    & Human3.6M \\ 
    \midrule
    \multicolumn{2}{c}{\textbf{\emph{fully supervised}}}              \\
    Newell~\etal~\cite{newell2016stacked}                     & 19.52 \\ 
    \midrule
    \multicolumn{2}{c}{\textbf{\emph{self-supervised + supervised post-processing}}}           \\
    Jakab \& Gupta~\etal~\cite{jakabunsupervised}                      & 19.12  \\
    \midrule
    \multicolumn{2}{c}{\textbf{\emph{self-supervised (no post-processing)}}}          \\
    KeypointGAN with \textit{3DHP prior} (ours)                   & 18.94    \\
    KeypointGAN with \textit{H3.6M prior} (ours)                  & \textbf{14.46} \\
    \bottomrule
    \end{tabular}%
  \smallskip
  \caption{{\bf Human landmark detection (full H3.6M).} 
  Comparison on Human3.6M test set with a supervised baseline Newell~\etal~\cite{newell2016stacked}, and a self-supervised method~\cite{jakabunsupervised}. 
  We report the MSE in pixels~\cite{h36m_pami}.
  Results for each activity are in the supplementary.}
  \label{tab:pose-sota}
\end{table}

\subsection{Human faces}\label{s:faces}

\paragraph{Datasets.}

\emph{VoxCeleb2~\cite{chung2018voxceleb2}} is a large-scale dataset consisting of 1M short clips of talking-head videos extracted from YouTube.
\emph{MultiPIE~\cite{sim2002cmu}} contains 68 labelled facial landmarks and 6k samples.
We use this dataset as the only source for the prior.
\emph{300-W~\cite{sagonas2016300}} is a challenging dataset of facial images obtained by combining multiple datasets~\cite{belhumeur2013localizing,zhou2013extensive,ramanan2012face} as described in~\cite{Thewlis17,ren2014face}.
As in \mbox{MultiPIE}, 300-W contains 68 annotated facial landmarks.
We use 300-W as our test dataset and follow the evaluation protocol in \cite{ren2014face}.

\begin{figure}[h]
\includegraphics[height=0.192\linewidth,width=0.192\linewidth]{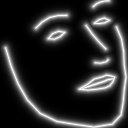}
\includegraphics[height=0.192\linewidth,width=0.192\linewidth]{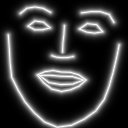}
\includegraphics[height=0.192\linewidth,width=0.192\linewidth]{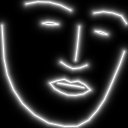}
\includegraphics[height=0.192\linewidth,width=0.192\linewidth]{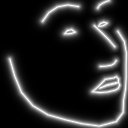}
\includegraphics[height=0.192\linewidth,width=0.192\linewidth]{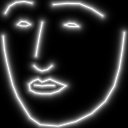}
\\
\includegraphics[height=0.192\linewidth,width=0.192\linewidth]{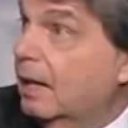}
\includegraphics[height=0.192\linewidth,width=0.192\linewidth]{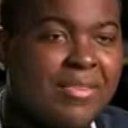}
\includegraphics[height=0.192\linewidth,width=0.192\linewidth]{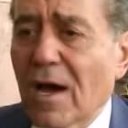}
\includegraphics[height=0.192\linewidth,width=0.192\linewidth]{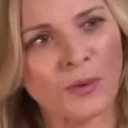}
\includegraphics[height=0.192\linewidth,width=0.192\linewidth]{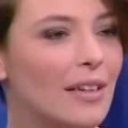}
\\
\includegraphics[height=0.192\linewidth,width=0.192\linewidth]{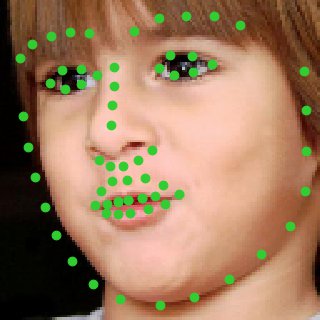}
\includegraphics[height=0.192\linewidth,width=0.192\linewidth]{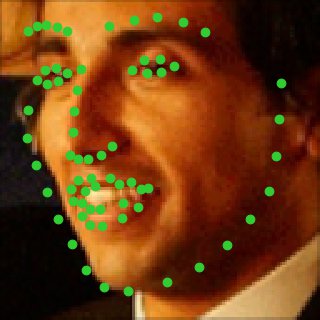}
\includegraphics[height=0.192\linewidth,width=0.192\linewidth]{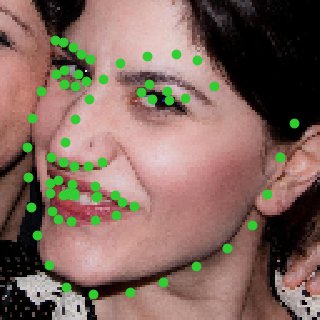}
\includegraphics[height=0.192\linewidth,width=0.192\linewidth]{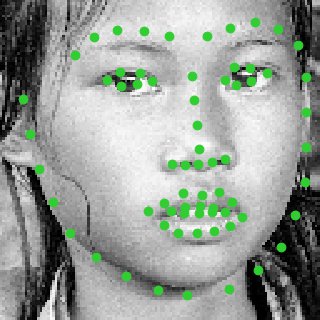}
\includegraphics[height=0.192\linewidth,width=0.192\linewidth]{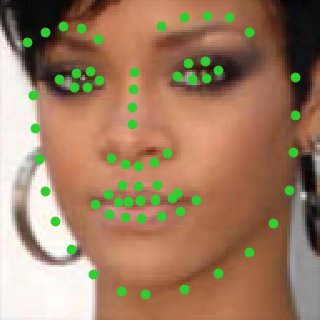}
\caption{\textbf{Unpaired transfer.} %
We leverage approx. 6k landmarks from the MultiPIE dataset~\cite{sim2002cmu} as a prior \textbf{[top]} and unlabelled images from the the large-scale VoxCeleb2~\cite{chung2018voxceleb2} \textbf{[middle]} (1M clips, 6k identities) to train a detector that we test on the 300-W dataset~\cite{sagonas2016300} \textbf{[bottom]} (predictions in green)
with state-of-the-art results (\cref{tab:face-sota}). More qualitative results are in the supplementary.}\label{f:transfer}
\end{figure}

\paragraph{Results.}
As for human pose, we study a scenario where images and poses are sourced from a \emph{different} datasets, using VoxCeleb2 and 300-W for the images, and MultiPIE (6k samples) for the poses~(\cref{f:transfer}).
We train KeypointGAN from scratch using video frames from VoxCeleb2; 
then we fine-tune the model using our unsupervised method on the 300-W training images. %
We report performance on 300-W test set in~\cref{tab:face-sota}.
KeypointGAN performs well even without any supervised fine-tuning on the target 300-W, 
and it already outperforms the unsupervised method of~\cite{Thewlis17a}.
Adding supervised post-processing (on 300-W training set) as done in all self-supervised learning methods~\cite{Thewlis17a,Thewlis17, Wiles18a,thewlis2019unsupervised}, we outperform all except for~\cite{thewlis2019unsupervised} when they use their \emph{HG} network that has 3 times more learnable parameters (4M vs 12M parameters).
Interestingly we also outperform all supervised methods except~\cite{xiao2016robust,feng2018wing}.

\begin{table}[t]
  \centering
  \begin{tabular}{@{}lr@{}}
    \toprule
    Method                                                    & 300-W \\ \midrule
    \multicolumn{2}{c}{\textbf{\emph{fully supervised}}}                                    \\
    LBF~\cite{ren2014face}                                    & 6.32  \\
    CFSS~\cite{zhu2015face}                                   & 5.76  \\
    cGPRT~\cite{lee2015face}                                             & 5.71  \\
    DDN~\cite{yu2016deep}                                               & 5.65  \\
    TCDCN~\cite{zhang2016learning}                                             & 5.54  \\
    RAR~\cite{xiao2016robust}                                               & 4.94  \\
    Wing Loss~\cite{feng2018wing}                                               & \textbf{4.04}  \\ \midrule
    \multicolumn{2}{c}{\textbf{\emph{self-supervised + supervised post-processing}}}           \\
    Thewlis~\etal~\cite{Thewlis17a}                           & 9.30  \\
    Thewlis~\etal~\cite{Thewlis17}                            & 7.97  \\
    Thewlis~\etal~\cite{thewlis2019unsupervised} \emph{SmallNet} $^\dagger$   & 5.75  \\
    Wiles~\etal~\cite{Wiles18a}                               & 5.71  \\
    Jakab \& Gupta~\etal~\cite{jakabunsupervised}                               & 5.39  \\
    Thewlis~\etal~\cite{thewlis2019unsupervised} \emph{HourGlass} $^\dagger$     & \textbf{4.65}  \\
    \midrule
    \multicolumn{2}{c}{\textbf{\emph{self-supervised}}}           \\
    \textbf{KeypointGAN} (ours w/o post-processing)                   & 8.67  \\
    \hspace{2mm} \textbf{+ supervised post-processing}                             & \textbf{5.12} \\
    \bottomrule
    \end{tabular}
  \smallskip
  \caption{{\bf Facial landmark detection.}
  Comparison with state-of-the-art methods on 2D facial landmark detection. We report the inter-ocular distance normalised keypoint localisation error~\cite{zhang2016learning} (in $\%$; $\downarrow$ is better) on the 300-W test set.
  $\dagger$: \cite{thewlis2019unsupervised} evaluate using two different networks: (1) \emph{SmallNet} which we outperform, (2) \emph{HourGlass} is not directly comparable due to much larger capacity (4M vs 12M parameters).}
  \label{tab:face-sota}
\end{table}

\subsection{Cat heads}\label{s:animals}

\begin{figure}[h]
\includegraphics[height=0.192\linewidth,width=0.192\linewidth]{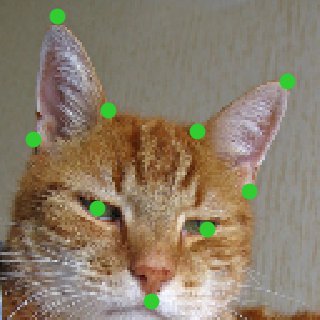}
\includegraphics[height=0.192\linewidth,width=0.192\linewidth]{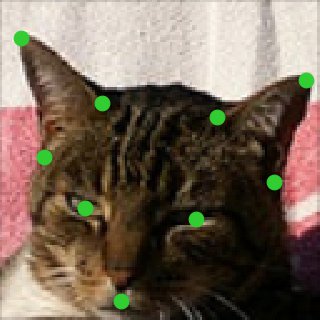}
\includegraphics[height=0.192\linewidth,width=0.192\linewidth]{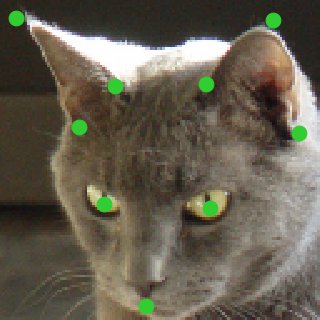}
\includegraphics[height=0.192\linewidth,width=0.192\linewidth]{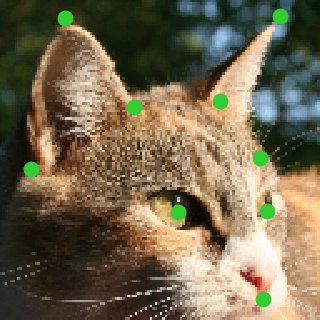}
\includegraphics[height=0.192\linewidth,width=0.192\linewidth]{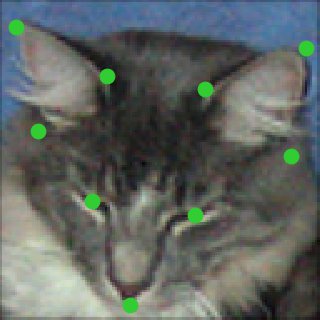}
\caption{\textbf{Cat head landmarks.}
Our predictions on Cat Head test set~\cite{zhang2008cat} consistently track landmarks across different views.
More results are included in the supplementary.
}\label{f:animals}
\end{figure}

\emph{Cat Head}~\cite{zhang2008cat} dataset contains 9k images of cat heads each annotated with 7 landmarks.
We use the same train and test split as~\cite{zhang2018unsupervised}.
We split the training set into two equally sized parts with no overlap.
The first one is used to sample training images and the second one for the landmark prior.
Our predictions are visualized in~\cref{f:animals}.

\subsection{Ablation study}\label{s:x-abal}

As noted above, we can obtain our method by making the following changes to CycleGAN:\@
(1)~switching to a conditional image generator $\Psi$,
(2)~introducing the skeleton bottleneck~$\beta\circ\eta$, and
(3)~removing the ``second auto-encoder cycle'' for the other domain (in our case the skeleton images).
\cref{tab:abal} shows the effect of modifying CycleGAN in this manner on Simplified Human3.6M~\cite{zhang2018unsupervised} for humans and on 300-W~\cite{sagonas2016300} for faces.

The baseline CycleGAN can be thought of as learning a mapping between images and skeletons via off-the-shelf image translation.
Switching to a conditional image generator (1) does not improve the results because the model can still leak appearance information's pose.
However, introducing the bottleneck (2) improves performance significantly for both humans ($2.86\%$ \emph{vs.} $3.54\%$ CycleGAN, a 20\% error reduction) and faces ($11.89\%$ \emph{vs.} $9.64\%$ CycleGAN, a 19\% error reduction).
This also justifies the use of a conditional generator as the model fails to converge if the bottleneck is used without it.
Removing the second cycle (3) leads to further improvements, showing that this part is detrimental for our task.

\begin{table}[th]
  \setlength{\tabcolsep}{2pt}
  \centering
  \begin{tabular}{@{}lrr@{}}
    \toprule
    Method                                  & humans & faces  \\ \midrule
    CycleGAN                                & 3.54   &  11.89 \\
    \hspace{1.5mm} + conditional generator (1)    & 3.60   &  --    \\
    \hspace{3mm} + skeleton-bottleneck (2)     & 2.86   &  9.64  \\
    \hspace{4.5mm} $-$ $2^{\text{nd}}$ cycle = \textbf{KeypointGAN} (ours) (3) & 2.73  & 8.67   \\ 
    CycleGAN $-$ $2^{\text{nd}}$ cycle  & 3.39   &  11.36 \\ \bottomrule
    \end{tabular}
  \smallskip
  \caption{{\bf Ablation study.} We start with the CycleGAN~\cite{zhu2017unpaired} model and sequentially augment it with --- (1) conditional image generator~($\Psi$), (2) skeleton bottleneck~($\beta\circ\eta$),
  and (3) remove the second cycle-constraint resulting in our proposed KeypointGAN model.
  An auto-encoding model with a skeleton image as the intermediate representation (\ie no keypoint bottleneck)
  and an adversarial loss is also reported (last row).
  We report 2D landmark detection error ($\downarrow$ is better) on the Simplified Human3.6M (\cref{s:x-human}) for human pose, on the 300-W (\cref{s:faces}) for faces.
  }
  \label{tab:abal}
\end{table}

\subsection{Unpaired sample efficiency}\label{s:x-sample}
\Cref{tab:mpie} demonstrates that KeypointGAN retains state-of-the-art performance even when we use only 50 unpaired landmark samples for the empirical prior.
The experiment was done following the same protocols for training on face and human datasets as described previously.
\begin{table}[th]
  \setlength{\tabcolsep}{2.2pt}
  \centering
  \begin{tabular}{@{}llll@{}}
    \toprule
    \multicolumn{1}{l|}{\# unpaired} & \multicolumn{1}{l|}{humans}               & \multicolumn{2}{c}{faces}  \\ 
    \multicolumn{1}{l|}{samples}     & \multicolumn{1}{l|}{\emph{no post-proc.}} & \emph{no post-proc.} &  \emph{+ sup. post-proc.} \\
    \midrule
    \multicolumn{1}{l|}{full dataset}                            & \multicolumn{1}{l|}{$2.73$}          & $8.67$          & $5.12$ \\
    \multicolumn{1}{l|}{5000}                                    & \multicolumn{1}{l|}{$2.92 \pm 0.05$} & --              & -- \\
    \multicolumn{1}{l|}{500}                                     & \multicolumn{1}{l|}{$3.30 \pm 0.06$} & $8.91 \pm 0.15$ & $5.22 \pm 0.04$ \\
    \multicolumn{1}{l|}{50}                                      & \multicolumn{1}{l|}{$4.05 \pm 0.02$} & $8.92 \pm 0.20$ & $5.19 \pm 0.06$ \\
    \bottomrule
    \end{tabular}
  \smallskip

  \caption{{\bf Varying \# of unpaired landmark samples.} 
  We train KeypointGAN using varying numbers of samples for landmark prior.
  For faces, we sample the prior from MultiPIE dataset and evaluate on 300-W (\cref{s:faces}).
  For human pose, we sample the prior from the disjoint part of the Simplified Human3.6M training set and evaluate on the test set (\cref{s:x-human}).
  We report the keypoint localisation error ($\pm \sigma$) (in $\%$; $\downarrow$ is better).
  Full dataset has 6k unpaired samples for faces, and 400k for humans.
  Decreasing the number of unpaired landmark samples retains most of the performance.
  }
  \label{tab:mpie}
\end{table}

\subsection{Appearance and geometry factorization}\label{s:factor}

\begin{figure}[t]
  \centering
  \begin{minipage}{\textwidth}
    \begin{minipage}[b]{4em}
    {\footnotesize
    \rotatebox{90}{\hspace{-2mm}target $\bx$}\\
    }
    \end{minipage}\hspace{-11mm}%
    \includegraphics[width=0.072\textwidth]{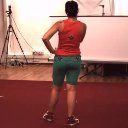}
    \includegraphics[width=0.072\textwidth]{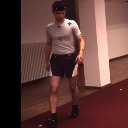}
    \includegraphics[width=0.072\textwidth]{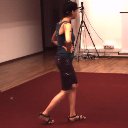}
    \includegraphics[width=0.072\textwidth]{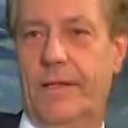}
    \includegraphics[width=0.072\textwidth]{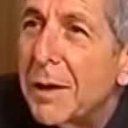}
    \includegraphics[width=0.072\textwidth]{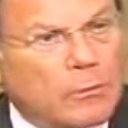}
    \\
    \begin{minipage}[b]{4em}
    {\footnotesize
    \rotatebox{90}{\hspace{-1mm}style $\bx'$}\\
    }
    \end{minipage}\hspace{-11mm}%
    \includegraphics[width=0.072\textwidth]{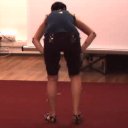}
    \includegraphics[width=0.072\textwidth]{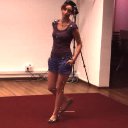}
    \includegraphics[width=0.072\textwidth]{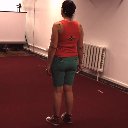}
    \includegraphics[width=0.072\textwidth]{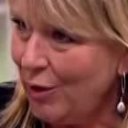}
    \includegraphics[width=0.072\textwidth]{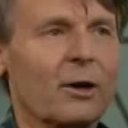}
    \includegraphics[width=0.072\textwidth]{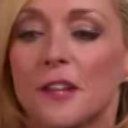}
    \\
    \begin{minipage}[t]{4em}
    {\footnotesize
    \rotatebox{90}{\hspace{3mm}rec.~$\hat\bx$}\\
    }
    \end{minipage}\hspace{-11mm}%
    \includegraphics[width=0.072\textwidth]{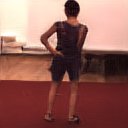}
    \includegraphics[width=0.072\textwidth]{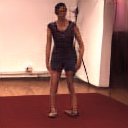}
    \includegraphics[width=0.072\textwidth]{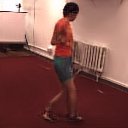}
    \includegraphics[width=0.072\textwidth]{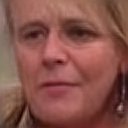}
    \includegraphics[width=0.072\textwidth]{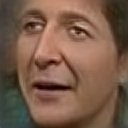}
    \includegraphics[width=0.072\textwidth]{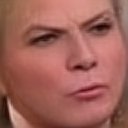}
    \\
  \end{minipage}\vspace{-1em}
  \caption{\textbf{Factorization of appearance and geometry.}
  \emph{Reconstructed} image inherits appearance from the \emph{style} image and geometry from the \emph{target} image.
  {\bf [left]:} human pose samples from Human3.6M.
  {\bf [right]:} face samples from VoxCeleb2.}\label{f:swap}
\end{figure}

The conditional image generator $\Psi: (\by^*, \bx') \mapsto \hat\bx$ of~\cref{e:autoencode} can also be used to produce novel images by combining pose and appearance from different images.
\Cref{f:swap} shows that the model can be used to transfer the appearance of a human face identity on top of the pose of another.
Though generating high quality images is not our primary goal, the ability to transfer appearance shows that KeypointGAN properly factorizes the latter from pose.

\begin{figure}[h]
  \centering
  \begin{minipage}{\textwidth}
    \begin{minipage}[b]{4em}
    {\footnotesize
    \rotatebox{90}{\hspace{0mm}}\\
    }
    \end{minipage}\hspace{-11mm}%
    \hspace{0.0840\textwidth}
    \begin{minipage}[b]{4em}
    {\footnotesize
    \rotatebox{90}{\hspace{-1mm}kpts $\bp$}\\
    }
    \end{minipage}\hspace{-11mm}%
    \includegraphics[width=0.084\textwidth]{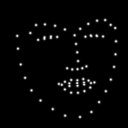}
    \includegraphics[width=0.084\textwidth]{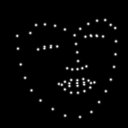}
    \includegraphics[width=0.084\textwidth]{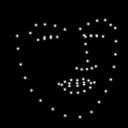}
    \includegraphics[width=0.084\textwidth]{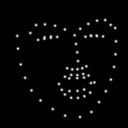}
    \\
    \begin{minipage}[b]{4em}
    {\footnotesize
    \rotatebox{90}{\hspace{-1mm}input $\bx$}\\
    }
    \end{minipage}\hspace{-11mm}%
    \includegraphics[width=0.0840\textwidth]{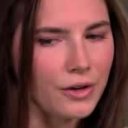}
    \begin{minipage}[b]{4em}
    {\footnotesize
    \rotatebox{90}{\hspace{-3mm}recons. $\hat\bx$}\\
    }
    \end{minipage}\hspace{-11mm}%
    \includegraphics[width=0.084\textwidth]{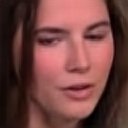}
    \includegraphics[width=0.084\textwidth]{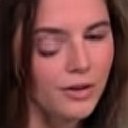}
    \includegraphics[width=0.084\textwidth]{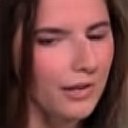}
    \includegraphics[width=0.084\textwidth]{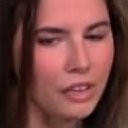}
    \\
    \hspace*{1.0000em}
    \hspace{0.1100\textwidth}
    \hspace*{1.0000em}
    \vspace{-4.0000mm}
    \end{minipage}
    \begin{minipage}{\textwidth}
    \begin{minipage}[b]{4em}
    {\footnotesize
    \rotatebox{90}{\hspace{0mm}}\\
    }
    \end{minipage}\hspace{-11mm}%
    \hspace{0.0840\textwidth}
    \begin{minipage}[b]{4em}
    {\footnotesize
    \rotatebox{90}{\hspace{-1mm}kpts $\bp$}\\
    }
    \end{minipage}\hspace{-11mm}%
    \includegraphics[width=0.084\textwidth]{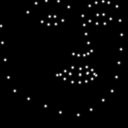}
    \includegraphics[width=0.084\textwidth]{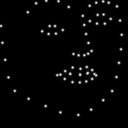}
    \includegraphics[width=0.084\textwidth]{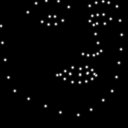}
    \includegraphics[width=0.084\textwidth]{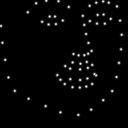}
    \\
    \begin{minipage}[b]{4em}
    {\footnotesize
    \rotatebox{90}{\hspace{-2mm}input $\bx$}\\
    }
    \end{minipage}\hspace{-11mm}%
    \includegraphics[width=0.0840\textwidth]{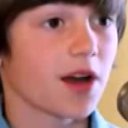}
    \begin{minipage}[b]{4em}
    {\footnotesize
    \rotatebox{90}{\hspace{-3mm}recons. $\hat\bx$}\\
    }
    \end{minipage}\hspace{-11mm}%
    \includegraphics[width=0.084\textwidth]{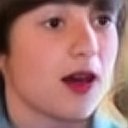}
    \includegraphics[width=0.084\textwidth]{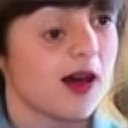}
    \includegraphics[width=0.084\textwidth]{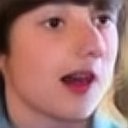}
    \includegraphics[width=0.084\textwidth]{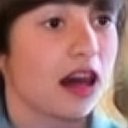}
    \\
    \hspace*{5.5em}
    {\footnotesize
    \begin{tabu} to 0.37\textwidth{X[c] X[c] X[c] X[c]}
    \vspace{-3.5000mm}original & \vspace{-3.5000mm}eye & \vspace{-3.5000mm}nose & \vspace{-3.5000mm}mouth
    \end{tabu}
    }
    \vspace{0.0000mm}
  \end{minipage}
  \caption{\textbf{Image editing using detected landmarks.}
  We show fine-grained control over the generated image by manipulating the coordinates of detected keypoints (\emph{kpts}).
  The resulting changes are localised.
  Apart from demonstrating successful disentanglement of appearance and geometry, this also suggests that KeypointGAN assigns correct semantics to the detected landmarks.
  }
  \label{f:move_face}
\end{figure}

This also demonstrates significant generalization over the training setting, as the system only learns from pairs of frames sampled from the same video and thus with same identity, but it can swap different identities.
In~\cref{f:move_face}, we further leverage the disentanglement of geometry and appearance to manipulate a face by editing its keypoints.

\section{Conclusion}\label{s:conc}

We have shown that combining conditional image generation with a dual representation of pose with a tight geometric bottleneck can be used to learn to recognize the pose of complex objects such as humans without providing any labelled image to the system.
In order to do so, KeypointGAN makes use of an unpaired pose prior, which also allows it to output human-interpretable pose parameters.
With this, we have achieved optimal landmark detection accuracy for methods that do not use labelled images for training.

\paragraph{Acknowledgements.}
We are grateful for the support of ERC 638009-IDIU, and the Clarendon Fund Scholarship.
We would like to thank Triantafyllos Afouras, Relja \mbox{Arandjelović}, and Chuhan Zhang for helpful advice.

\let\oldthebibliography\thebibliography
\let\endoldthebibliography\endthebibliography
\renewenvironment{thebibliography}[1]{
  \begin{oldthebibliography}{#1}
    \setlength{\itemsep}{0em}
    \setlength{\parskip}{0em}
}
{
  \end{oldthebibliography}
}
{\small\bibliographystyle{plainnat}\bibliography{longstrings,refs}}

\begin{thebibliography}{82}
\providecommand{\natexlab}[1]{#1}
\providecommand{\url}[1]{\texttt{#1}}
\expandafter\ifx\csname urlstyle\endcsname\relax
  \providecommand{\doi}[1]{doi: #1}\else
  \providecommand{\doi}{doi: \begingroup \urlstyle{rm}\Url}\fi

\bibitem[Andriluka et~al.(2009)Andriluka, Roth, and
  Schiele]{andriluka2009pictorial}
Mykhaylo Andriluka, Stefan Roth, and Bernt Schiele.
\newblock Pictorial structures revisited: People detection and articulated pose
  estimation.
\newblock In \emph{Proc. CVPR}, pages 1014--1021. IEEE, 2009.

\bibitem[Andriluka et~al.(2014)Andriluka, Pishchulin, Gehler, and
  Schiele]{andriluka20142d}
Mykhaylo Andriluka, Leonid Pishchulin, Peter Gehler, and Bernt Schiele.
\newblock 2d human pose estimation: New benchmark and state of the art
  analysis.
\newblock In \emph{Proc. CVPR}, pages 3686--3693, 2014.

\bibitem[Belagiannis and Zisserman(2017)]{belagiannis2017recurrent}
Vasileios Belagiannis and Andrew Zisserman.
\newblock Recurrent human pose estimation.
\newblock In \emph{2017 12th IEEE International Conference on Automatic Face \&
  Gesture Recognition (FG 2017)}, pages 468--475. IEEE, 2017.

\bibitem[Belhumeur et~al.(2013)Belhumeur, Jacobs, Kriegman, and
  Kumar]{belhumeur2013localizing}
Peter~N Belhumeur, David~W Jacobs, David~J Kriegman, and Neeraj Kumar.
\newblock Localizing parts of faces using a consensus of exemplars.
\newblock \emph{TPAMI}, 35\penalty0 (12):\penalty0 2930--2940, 2013.

\bibitem[Bruna et~al.(2016)Bruna, Sprechmann, and LeCun]{bruna2015super}
Joan Bruna, Pablo Sprechmann, and Yann LeCun.
\newblock Super-resolution with deep convolutional sufficient statistics.
\newblock In \emph{Proc. ICLR}, 2016.

\bibitem[Bulat and Tzimiropoulos(2016)]{bulat2016human}
Adrian Bulat and Georgios Tzimiropoulos.
\newblock Human pose estimation via convolutional part heatmap regression.
\newblock In \emph{Proc. ECCV}, pages 717--732. Springer, 2016.

\bibitem[Cao et~al.(2017)Cao, Simon, Wei, and Sheikh]{cao2017realtime}
Zhe Cao, Tomas Simon, Shih-En Wei, and Yaser Sheikh.
\newblock Realtime multi-person 2d pose estimation using part affinity fields.
\newblock In \emph{Proc. CVPR}, pages 7291--7299, 2017.

\bibitem[Carreira et~al.(2016)Carreira, Agrawal, Fragkiadaki, and
  Malik]{carreira2016human}
Joao Carreira, Pulkit Agrawal, Katerina Fragkiadaki, and Jitendra Malik.
\newblock Human pose estimation with iterative error feedback.
\newblock In \emph{Proc. CVPR}, pages 4733--4742, 2016.

\bibitem[Chen and Yuille(2014)]{chen2014articulated}
Xianjie Chen and Alan~L Yuille.
\newblock Articulated pose estimation by a graphical model with image dependent
  pairwise relations.
\newblock In \emph{Proc. NIPS}, pages 1736--1744, 2014.

\bibitem[Chung et~al.(2018)Chung, Nagrani, and Zisserman]{chung2018voxceleb2}
Joon~Son Chung, Arsha Nagrani, and Andrew Zisserman.
\newblock Voxceleb2: Deep speaker recognition.
\newblock \emph{arXiv preprint arXiv:1806.05622}, 2018.

\bibitem[Dosovitskiy and Brox(2016)]{dosovitskiy2016generating}
Alexey Dosovitskiy and Thomas Brox.
\newblock Generating images with perceptual similarity metrics based on deep
  networks.
\newblock In \emph{Advances in Neural Information Processing Systems}, pages
  658--666, 2016.

\bibitem[Felzenszwalb and Huttenlocher(2005)]{felzenszwalb2005pictorial}
Pedro~F Felzenszwalb and Daniel~P Huttenlocher.
\newblock Pictorial structures for object recognition.
\newblock \emph{International journal of computer vision}, 61\penalty0
  (1):\penalty0 55--79, 2005.

\bibitem[Feng et~al.(2018)Feng, Kittler, Awais, Huber, and Wu]{feng2018wing}
Zhen-Hua Feng, Josef Kittler, Muhammad Awais, Patrik Huber, and Xiao-Jun Wu.
\newblock Wing loss for robust facial landmark localisation with convolutional
  neural networks.
\newblock In \emph{Proc. CVPR}, 2018.

\bibitem[Ganin and Lempitsky(2015)]{ganin2015unsupervised}
Yaroslav Ganin and Victor Lempitsky.
\newblock Unsupervised domain adaptation by backpropagation.
\newblock In \emph{International Conference on Machine Learning}, pages
  1180--1189, 2015.

\bibitem[Gatys et~al.(2016)Gatys, Ecker, and Bethge]{gatys2016image}
Leon~A Gatys, Alexander~S Ecker, and Matthias Bethge.
\newblock Image style transfer using convolutional neural networks.
\newblock In \emph{Proc. CVPR}, pages 2414--2423, 2016.

\bibitem[Gecer et~al.(2019)Gecer, Ploumpis, Kotsia, and
  Zafeiriou]{gecer19ganfit}
Baris Gecer, Stylianos Ploumpis, Irene Kotsia, and Stefanos Zafeiriou.
\newblock {GANFIT}: Generative adversarial network fitting for high fidelity
  {3D} face reconstruction.
\newblock In \emph{Proceedings of the IEEE Conference on Computer Vision and
  Pattern Recognition}, 2019.

\bibitem[Geng et~al.(2019)Geng, Cao, and Tulyakov]{geng193d-guided}
Zhenglin Geng, Chen Cao, and Sergey Tulyakov.
\newblock {3D} guided fine-grained face manipulation.
\newblock In \emph{Proceedings of the IEEE Conference on Computer Vision and
  Pattern Recognition}, 2019.

\bibitem[Gerig et~al.(2018)Gerig, Morel-Forster, Blumer, Egger, Luthi,
  Sch{\"o}nborn, and Vetter]{gerig18morphable}
Thomas Gerig, Andreas Morel-Forster, Clemens Blumer, Bernhard Egger, Marcel
  Luthi, Sandro Sch{\"o}nborn, and Thomas Vetter.
\newblock Morphable face models - an open framework.
\newblock In \emph{Proc. Automatic Face \& Gesture Recognition}, 2018.

\bibitem[Goodfellow et~al.(2014)Goodfellow, Pouget-Abadie, Mirza, Xu,
  Warde-Farley, Ozair, Courville, and Bengio]{goodfellow2014generative}
Ian Goodfellow, Jean Pouget-Abadie, Mehdi Mirza, Bing Xu, David Warde-Farley,
  Sherjil Ozair, Aaron Courville, and Yoshua Bengio.
\newblock Generative adversarial nets.
\newblock In \emph{Proc. NIPS}, pages 2672--2680, 2014.

\bibitem[Gupta et~al.(2018)Gupta, Vedaldi, and Zisserman]{gupta2018learning}
Ankush Gupta, Andrea Vedaldi, and Andrew Zisserman.
\newblock Learning to read by spelling: Towards unsupervised text recognition.
\newblock In \emph{Proc. ICVGIP}, 2018.

\bibitem[Hoffman et~al.(2017)Hoffman, Tzeng, Park, Zhu, Isola, Saenko, Efros,
  and Darrell]{hoffman2017cycada}
Judy Hoffman, Eric Tzeng, Taesung Park, Jun-Yan Zhu, Phillip Isola, Kate
  Saenko, Alexei~A Efros, and Trevor Darrell.
\newblock Cycada: Cycle-consistent adversarial domain adaptation.
\newblock \emph{arXiv preprint arXiv:1711.03213}, 2017.

\bibitem[Insafutdinov et~al.(2016)Insafutdinov, Pishchulin, Andres, Andriluka,
  and Schiele]{insafutdinov2016deepercut}
Eldar Insafutdinov, Leonid Pishchulin, Bjoern Andres, Mykhaylo Andriluka, and
  Bernt Schiele.
\newblock Deepercut: A deeper, stronger, and faster multi-person pose
  estimation model.
\newblock In \emph{Proc. ECCV}, pages 34--50. Springer, 2016.

\bibitem[Ionescu et~al.(2014)Ionescu, Papava, Olaru, and
  Sminchisescu]{h36m_pami}
Catalin Ionescu, Dragos Papava, Vlad Olaru, and Cristian Sminchisescu.
\newblock Human3.6m: Large scale datasets and predictive methods for 3d human
  sensing in natural environments.
\newblock \emph{TPAMI}, 36\penalty0 (7):\penalty0 1325--1339, jul 2014.

\bibitem[Isola et~al.(2017)Isola, Zhu, Zhou, and Efros]{isola2017image}
Phillip Isola, Jun-Yan Zhu, Tinghui Zhou, and Alexei~A Efros.
\newblock Image-to-image translation with conditional adversarial networks.
\newblock In \emph{Proc. CVPR}, 2017.

\bibitem[Jakab et~al.(2018)Jakab, Gupta, Bilen, and Vedaldi]{jakabunsupervised}
Tomas Jakab, Ankush Gupta, Hakan Bilen, and Andrea Vedaldi.
\newblock Unsupervised learning of object landmarks through conditional image
  generation.
\newblock In \emph{Proc. NIPS}, 2018.

\bibitem[Johnson et~al.(2016)Johnson, Alahi, and
  Fei-Fei]{johnson2016perceptual}
Justin Johnson, Alexandre Alahi, and Li~Fei-Fei.
\newblock Perceptual losses for real-time style transfer and super-resolution.
\newblock In \emph{Proc. ECCV}, pages 694--711. Springer, 2016.

\bibitem[Johnson and Everingham(2011)]{johnson2011learning}
Sam Johnson and Mark Everingham.
\newblock Learning effective human pose estimation from inaccurate annotation.
\newblock In \emph{Proc. CVPR}, pages 1465--1472. IEEE, 2011.

\bibitem[Kanazawa et~al.(2016)Kanazawa, Jacobs, and
  Chandraker]{kanazawa2016warpnet}
Angjoo Kanazawa, David~W Jacobs, and Manmohan Chandraker.
\newblock Warpnet: Weakly supervised matching for single-view reconstruction.
\newblock In \emph{Proc. CVPR}, pages 3253--3261, 2016.

\bibitem[Kanazawa et~al.(2018)Kanazawa, Black, Jacobs, and
  Malik]{kanazawa2018end}
Angjoo Kanazawa, Michael~J Black, David~W Jacobs, and Jitendra Malik.
\newblock End-to-end recovery of human shape and pose.
\newblock In \emph{Proc. CVPR}, 2018.

\bibitem[Karras et~al.(2017)Karras, Aila, Laine, and
  Lehtinen]{karras2017progressive}
Tero Karras, Timo Aila, Samuli Laine, and Jaakko Lehtinen.
\newblock Progressive growing of gans for improved quality, stability, and
  variation.
\newblock \emph{arXiv preprint arXiv:1710.10196}, 2017.

\bibitem[Kingma and Ba(2014)]{kingma2014adam}
Diederik~P Kingma and Jimmy Ba.
\newblock Adam: A method for stochastic optimization.
\newblock \emph{arXiv preprint arXiv:1412.6980}, 2014.

\bibitem[Lee et~al.(2015)Lee, Park, and Yoo]{lee2015face}
Donghoon Lee, Hyunsin Park, and Chang~D Yoo.
\newblock Face alignment using cascade gaussian process regression trees.
\newblock In \emph{Proc. CVPR}, 2015.

\bibitem[Lin et~al.(2014)Lin, Maire, Belongie, Hays, Perona, Ramanan,
  Doll{\'a}r, and Zitnick]{lin2014microsoft}
Tsung-Yi Lin, Michael Maire, Serge Belongie, James Hays, Pietro Perona, Deva
  Ramanan, Piotr Doll{\'a}r, and C~Lawrence Zitnick.
\newblock Microsoft coco: Common objects in context.
\newblock In \emph{Proc. ECCV}, pages 740--755. Springer, 2014.

\bibitem[Loper et~al.(2015)Loper, Mahmood, Romero, Pons-Moll, and
  Black]{loper2015smpl}
Matthew Loper, Naureen Mahmood, Javier Romero, Gerard Pons-Moll, and Michael~J
  Black.
\newblock {SMPL}: A skinned multi-person linear model.
\newblock \emph{ACM transactions on graphics (TOG)}, 34\penalty0 (6):\penalty0
  248, 2015.

\bibitem[Lorenz et~al.(2019)Lorenz, Bereska, Milbich, and
  Ommer]{lorenz2019unsupervised}
Dominik Lorenz, Leonard Bereska, Timo Milbich, and Bjorn Ommer.
\newblock Unsupervised part-based disentangling of object shape and appearance.
\newblock In \emph{Proceedings of the IEEE Conference on Computer Vision and
  Pattern Recognition}, pages 10955--10964, 2019.

\bibitem[Maas et~al.(2013)Maas, Hannun, and Ng]{maas2013rectifier}
Andrew~L Maas, Awni~Y Hannun, and Andrew~Y Ng.
\newblock Rectifier nonlinearities improve neural network acoustic models.
\newblock In \emph{Proc. icml}, volume~30, page~3, 2013.

\bibitem[Mao et~al.(2017)Mao, Li, Xie, Lau, Wang, and
  Paul~Smolley]{mao2017least}
Xudong Mao, Qing Li, Haoran Xie, Raymond~YK Lau, Zhen Wang, and Stephen
  Paul~Smolley.
\newblock Least squares generative adversarial networks.
\newblock In \emph{Proc. ICCV}, pages 2794--2802, 2017.

\bibitem[Mehta et~al.(2017)Mehta, Rhodin, Casas, Fua, Sotnychenko, Xu, and
  Theobalt]{mehta2017monocular}
Dushyant Mehta, Helge Rhodin, Dan Casas, Pascal Fua, Oleksandr Sotnychenko,
  Weipeng Xu, and Christian Theobalt.
\newblock Monocular 3d human pose estimation in the wild using improved cnn
  supervision.
\newblock In \emph{2017 International Conference on 3D Vision (3DV)}, pages
  506--516. IEEE, 2017.

\bibitem[Newell et~al.(2016)Newell, Yang, and Deng]{newell2016stacked}
Alejandro Newell, Kaiyu Yang, and Jia Deng.
\newblock Stacked hourglass networks for human pose estimation.
\newblock In \emph{Proc. ECCV}, 2016.

\bibitem[Ouyang et~al.(2014)Ouyang, Chu, and Wang]{ouyang2014multi}
Wanli Ouyang, Xiao Chu, and Xiaogang Wang.
\newblock Multi-source deep learning for human pose estimation.
\newblock In \emph{Proc. CVPR}, pages 2329--2336, 2014.

\bibitem[Paysan et~al.(2009)Paysan, Knothe, Amberg, Romdhani, and
  Vetter]{bfm09}
Pascal Paysan, Reinhard Knothe, Brian Amberg, Sami Romdhani, and Thomas Vetter.
\newblock A 3d face model for pose and illumination invariant face recognition.
\newblock In \emph{The IEEE International Conference on Advanced Video and
  Signal Based Surveillance}, 2009.

\bibitem[Pfister et~al.(2015)Pfister, Charles, and
  Zisserman]{pfister2015flowing}
Tomas Pfister, James Charles, and Andrew Zisserman.
\newblock Flowing convnets for human pose estimation in videos.
\newblock In \emph{Proc. CVPR}, pages 1913--1921, 2015.

\bibitem[Pishchulin et~al.(2013)Pishchulin, Andriluka, Gehler, and
  Schiele]{pishchulin2013poselet}
Leonid Pishchulin, Mykhaylo Andriluka, Peter Gehler, and Bernt Schiele.
\newblock Poselet conditioned pictorial structures.
\newblock In \emph{Proc. CVPR}, pages 588--595, 2013.

\bibitem[Ramakrishna et~al.(2014)Ramakrishna, Munoz, Hebert, Bagnell, and
  Sheikh]{ramakrishna2014pose}
Varun Ramakrishna, Daniel Munoz, Martial Hebert, James~Andrew Bagnell, and
  Yaser Sheikh.
\newblock Pose machines: Articulated pose estimation via inference machines.
\newblock In \emph{Proc. ECCV}, pages 33--47. Springer, 2014.

\bibitem[Ramanan and Zhu(2012)]{ramanan2012face}
Deva Ramanan and Xiangxin Zhu.
\newblock Face detection, pose estimation, and landmark localization in the
  wild.
\newblock In \emph{Proc. CVPR}, 2012.

\bibitem[Ren et~al.(2014)Ren, Cao, Wei, and Sun]{ren2014face}
Shaoqing Ren, Xudong Cao, Yichen Wei, and Jian Sun.
\newblock Face alignment at 3000 fps via regressing local binary features.
\newblock In \emph{Proc. CVPR}, 2014.

\bibitem[Rhodin et~al.(2018)Rhodin, Salzmann, and Fua]{rhodin2018unsupervised}
Helge Rhodin, Mathieu Salzmann, and Pascal Fua.
\newblock Unsupervised geometry-aware representation for 3d human pose
  estimation.
\newblock In \emph{Proceedings of the European Conference on Computer Vision
  (ECCV)}, pages 750--767, 2018.

\bibitem[Rocco et~al.(2017)Rocco, Arandjelovic, and
  Sivic]{rocco2017convolutional}
Ignacio Rocco, Relja Arandjelovic, and Josef Sivic.
\newblock Convolutional neural network architecture for geometric matching.
\newblock In \emph{Proc. CVPR}, volume~2, 2017.

\bibitem[Ronchi et~al.(2018)Ronchi, Mac~Aodha, Eng, and
  Perona]{relativeposeBMVC18}
Matteo~Ruggero Ronchi, Oisin Mac~Aodha, Robert Eng, and Pietro Perona.
\newblock It's all relative: Monocular 3d human pose estimation from weakly
  supervised data.
\newblock In \emph{BMVC}, 2018.

\bibitem[Sagonas et~al.(2016)Sagonas, Antonakos, Tzimiropoulos, Zafeiriou, and
  Pantic]{sagonas2016300}
Christos Sagonas, Epameinondas Antonakos, Georgios Tzimiropoulos, Stefanos
  Zafeiriou, and Maja Pantic.
\newblock 300 faces in-the-wild challenge: Database and results.
\newblock \emph{Image and vision computing}, 47:\penalty0 3--18, 2016.

\bibitem[Sapp et~al.(2010)Sapp, Jordan, and Taskar]{sapp2010adaptive}
Benjamin Sapp, Chris Jordan, and Ben Taskar.
\newblock Adaptive pose priors for pictorial structures.
\newblock In \emph{Proc. CVPR}, pages 422--429, 2010.

\bibitem[Sengupta et~al.(2018)Sengupta, Kanazawa, Castillo, and
  Jacobs]{Sengupta18sfsnet}
Soumyadip Sengupta, Angjoo Kanazawa, Carlos~D. Castillo, and David Jacobs.
\newblock Sfsnet: Learning shape, refectance and illuminance of faces in the
  wild.
\newblock In \emph{Proceedings of the IEEE Conference on Computer Vision and
  Pattern Recognition}, 2018.

\bibitem[Shu et~al.(2018)Shu, Sahasrabudhe, Guler, Samaras, Paragios, and
  Kokkinos]{Shu2018}
Zhixin Shu, Mihir Sahasrabudhe, Alp Guler, Dimitris Samaras, Nikos Paragios,
  and Iasonas Kokkinos.
\newblock Deforming autoencoders: Unsupervised disentangling of shape and
  appearance.
\newblock In \emph{Proceedings of the European Conference on Computer Vision},
  2018.

\bibitem[Sim et~al.(2002)Sim, Baker, and Bsat]{sim2002cmu}
Terence Sim, Simon Baker, and Maan Bsat.
\newblock The cmu pose, illumination, and expression (pie) database.
\newblock In \emph{Proceedings of Fifth IEEE International Conference on
  Automatic Face Gesture Recognition}, pages 53--58. IEEE, 2002.

\bibitem[Singh et~al.(2019)Singh, Ojha, and Lee]{singh2019finegan}
Krishna~Kumar Singh, Utkarsh Ojha, and Yong~Jae Lee.
\newblock Finegan: Unsupervised hierarchical disentanglement for fine-grained
  object generation and discovery.
\newblock In \emph{Proceedings of the IEEE Conference on Computer Vision and
  Pattern Recognition}, pages 6490--6499, 2019.

\bibitem[Sundermeyer et~al.(2018)Sundermeyer, Marton, Durner, Brucker, and
  Triebel]{Sundermeyer2018}
Martin Sundermeyer, Zoltan-Csaba Marton, Maximilian Durner, Manuel Brucker, and
  Rudolph Triebel.
\newblock Implicit 3d orientation learning for 6d object detection from rgb
  images.
\newblock In \emph{Proc. ECCV}, pages 712--729. Springer, 2018.

\bibitem[Thewlis et~al.(2017{\natexlab{a}})Thewlis, Bilen, and
  Vedaldi]{Thewlis17}
James Thewlis, Hakan Bilen, and Andrea Vedaldi.
\newblock Unsupervised learning of object landmarks by factorized spatial
  embeddings.
\newblock In \emph{Proc. ICCV}, 2017{\natexlab{a}}.

\bibitem[Thewlis et~al.(2017{\natexlab{b}})Thewlis, Bilen, and
  Vedaldi]{Thewlis17a}
James Thewlis, Hakan Bilen, and Andrea Vedaldi.
\newblock Unsupervised learning of object frames by dense equivariant image
  labelling.
\newblock In \emph{Proc. NIPS}, 2017{\natexlab{b}}.

\bibitem[Thewlis et~al.(2019)Thewlis, Albanie, Bilen, and
  Vedaldi]{thewlis2019unsupervised}
James Thewlis, Samuel Albanie, Hakan Bilen, and Andrea Vedaldi.
\newblock Unsupervised learning of landmarks by descriptor vector exchange.
\newblock In \emph{Proceedings of the IEEE International Conference on Computer
  Vision}, pages 6361--6371, 2019.

\bibitem[Tompson et~al.(2015)Tompson, Goroshin, Jain, LeCun, and
  Bregler]{tompson2015efficient}
Jonathan Tompson, Ross Goroshin, Arjun Jain, Yann LeCun, and Christoph Bregler.
\newblock Efficient object localization using convolutional networks.
\newblock In \emph{Proc. CVPR}, pages 648--656, 2015.

\bibitem[Tompson et~al.(2014)Tompson, Jain, LeCun, and
  Bregler]{tompson2014joint}
Jonathan~J Tompson, Arjun Jain, Yann LeCun, and Christoph Bregler.
\newblock Joint training of a convolutional network and a graphical model for
  human pose estimation.
\newblock In \emph{Advances in neural information processing systems}, pages
  1799--1807, 2014.

\bibitem[Toshev and Szegedy(2014)]{toshev2014deeppose}
Alexander Toshev and Christian Szegedy.
\newblock Deeppose: Human pose estimation via deep neural networks.
\newblock In \emph{Proc. CVPR}, pages 1653--1660, 2014.

\bibitem[Tran et~al.(2017)Tran, Yin, and Liu]{tran2017disentangled}
Luan Tran, Xi~Yin, and Xiaoming Liu.
\newblock Disentangled representation learning gan for pose-invariant face
  recognition.
\newblock In \emph{Proceedings of the IEEE conference on computer vision and
  pattern recognition}, pages 1415--1424, 2017.

\bibitem[Tung et~al.(2017)Tung, Harley, Seto, and
  Fragkiadaki]{tung2017adversarial}
Hsiao-Yu~Fish Tung, Adam~W Harley, William Seto, and Katerina Fragkiadaki.
\newblock Adversarial inverse graphics networks: Learning 2d-to-3d lifting and
  image-to-image translation from unpaired supervision.
\newblock In \emph{Proc. ICCV}, volume~2, 2017.

\bibitem[Tzeng et~al.(2015)Tzeng, Hoffman, Darrell, and
  Saenko]{tzeng2015simultaneous}
Eric Tzeng, Judy Hoffman, Trevor Darrell, and Kate Saenko.
\newblock Simultaneous deep transfer across domains and tasks.
\newblock In \emph{Proc. CVPR}, pages 4068--4076, 2015.

\bibitem[Tzeng et~al.(2017)Tzeng, Hoffman, Saenko, and
  Darrell]{tzeng2017adversarial}
Eric Tzeng, Judy Hoffman, Kate Saenko, and Trevor Darrell.
\newblock Adversarial discriminative domain adaptation.
\newblock In \emph{Proc. CVPR}, 2017.

\bibitem[Ulyanov et~al.(2016)Ulyanov, Vedaldi, and
  Lempitsky]{ulyanov2016instance}
Dmitry Ulyanov, Andrea Vedaldi, and Victor Lempitsky.
\newblock Instance normalization: The missing ingredient for fast stylization.
\newblock \emph{arXiv preprint arXiv:1607.08022}, 2016.

\bibitem[Villegas et~al.(2017)Villegas, Yang, Zou, Sohn, Lin, and
  Lee]{villegas2017learning}
Ruben Villegas, Jimei Yang, Yuliang Zou, Sungryull Sohn, Xunyu Lin, and Honglak
  Lee.
\newblock Learning to generate long-term future via hierarchical prediction.
\newblock In \emph{Proc. ICML}, 2017.

\bibitem[Wang et~al.(2019)Wang, Shu, Cheng, Panagakis, Samaras, and
  Zafeiriou]{wang2019adversarial}
Mengjiao Wang, Zhixin Shu, Shiyang Cheng, Yannis Panagakis, Dimitris Samaras,
  and Stefanos Zafeiriou.
\newblock An adversarial neuro-tensorial approach for learning disentangled
  representations.
\newblock \emph{International Journal of Computer Vision}, 2019.

\bibitem[Wei et~al.(2016)Wei, Ramakrishna, Kanade, and
  Sheikh]{wei2016convolutional}
Shih-En Wei, Varun Ramakrishna, Takeo Kanade, and Yaser Sheikh.
\newblock Convolutional pose machines.
\newblock In \emph{Proc. CVPR}, pages 4724--4732, 2016.

\bibitem[Wiles et~al.(2018)Wiles, Koepke, and Zisserman]{Wiles18a}
Olivia Wiles, A~Koepke, and Andrew Zisserman.
\newblock Self-supervised learning of a facial attribute embedding from video.
\newblock In \emph{Proc. BMVC}, 2018.

\bibitem[Xiao et~al.(2016)Xiao, Feng, Xing, Lai, Yan, and
  Kassim]{xiao2016robust}
Shengtao Xiao, Jiashi Feng, Junliang Xing, Hanjiang Lai, Shuicheng Yan, and
  Ashraf Kassim.
\newblock Robust facial landmark detection via recurrent attentive-refinement
  networks.
\newblock In \emph{Proc. ECCV}, 2016.

\bibitem[Yang et~al.(2018)Yang, Ouyang, Wang, Ren, Li, and Wang]{yang20183d}
Wei Yang, Wanli Ouyang, Xiaolong Wang, Jimmy Ren, Hongsheng Li, and Xiaogang
  Wang.
\newblock 3d human pose estimation in the wild by adversarial learning.
\newblock In \emph{Proc. CVPR}, volume~1, 2018.

\bibitem[Yang and Ramanan(2011)]{yang2011articulated}
Yi~Yang and Deva Ramanan.
\newblock Articulated pose estimation with flexible mixtures-of-parts.
\newblock In \emph{Proc. CVPR}, pages 1385--1392. IEEE, 2011.

\bibitem[Yu et~al.(2016)Yu, Zhou, and Chandraker]{yu2016deep}
Xiang Yu, Feng Zhou, and Manmohan Chandraker.
\newblock Deep deformation network for object landmark localization.
\newblock In \emph{Proc. ECCV}. Springer, 2016.

\bibitem[Zhang et~al.(2008)Zhang, Sun, and Tang]{zhang2008cat}
Weiwei Zhang, Jian Sun, and Xiaoou Tang.
\newblock Cat head detection-how to effectively exploit shape and texture
  features.
\newblock In \emph{European Conference on Computer Vision}, pages 802--816.
  Springer, 2008.

\bibitem[Zhang et~al.(2013)Zhang, Zhu, and Derpanis]{zhang2013actemes}
Weiyu Zhang, Menglong Zhu, and Konstantinos~G Derpanis.
\newblock From actemes to action: A strongly-supervised representation for
  detailed action understanding.
\newblock In \emph{Proceedings of the IEEE International Conference on Computer
  Vision}, pages 2248--2255, 2013.

\bibitem[Zhang et~al.(2018)Zhang, Guo, Jin, Luo, He, and
  Lee]{zhang2018unsupervised}
Yuting Zhang, Yijie Guo, Yixin Jin, Yijun Luo, Zhiyuan He, and Honglak Lee.
\newblock Unsupervised discovery of object landmarks as structural
  representations.
\newblock In \emph{Proc. CVPR}, pages 2694--2703, 2018.

\bibitem[Zhang et~al.(2016)Zhang, Luo, Loy, and Tang]{zhang2016learning}
Zhanpeng Zhang, Ping Luo, Chen~Change Loy, and Xiaoou Tang.
\newblock Learning deep representation for face alignment with auxiliary
  attributes.
\newblock \emph{TPAMI}, 38\penalty0 (5):\penalty0 918--930, 2016.

\bibitem[Zhou et~al.(2013)Zhou, Fan, Cao, Jiang, and Yin]{zhou2013extensive}
Erjin Zhou, Haoqiang Fan, Zhimin Cao, Yuning Jiang, and Qi~Yin.
\newblock Extensive facial landmark localization with coarse-to-fine
  convolutional network cascade.
\newblock In \emph{Proceedings of the IEEE International Conference on Computer
  Vision Workshops}, pages 386--391, 2013.

\bibitem[Zhu et~al.(2018)Zhu, Park, Isola, and Efros]{zhu2017unpaired}
Jun-Yan Zhu, Taesung Park, Phillip Isola, and Alexei~A Efros.
\newblock Unpaired image-to-image translation using cycle-consistent
  adversarial networks.
\newblock In \emph{Proc. CVPR}, 2018.

\bibitem[Zhu et~al.(2015)Zhu, Li, Change~Loy, and Tang]{zhu2015face}
Shizhan Zhu, Cheng Li, Chen Change~Loy, and Xiaoou Tang.
\newblock Face alignment by coarse-to-fine shape searching.
\newblock In \emph{Proc. CVPR}, 2015.

\end{thebibliography}

\clearpage
\onecolumn
\appendix
\section*{Appendix}
This supplementary material provides further technical details, illustrations and analysis.
We provide detailed quantitative evaluation on Human3.6M dataset (\cref{as:pose-sota}), extended versions of our qualitative results on factorization of appearance and geometry (\cref{as:factor}), facial landmarks detection (\cref{as:face_test}), human pose estimation (\cref{as:human_test}), and cat head landmarks detection (\cref{as:cat_test}).
Finally, we give further implementation details in \cref{as:architectures}.

\section{Human3.6M detailed results}\label{as:pose-sota}
We report the performance for each activity of the Human3.6M test set in \cref{atab:pose-sota}. We evaluated the performance on every 60th frame of the video sequences.
\begin{table}[h]
  \setlength{\tabcolsep}{2pt}
  \centering
  \resizebox{\textwidth}{!}{
    \begin{tabular}{@{}ll|r|ccccccccccccccc@{}}
      \toprule
                                                            & Method                                                      & \multicolumn{1}{c|}{all}               & \multicolumn{1}{c}{wait}               & \multicolumn{1}{c}{pose}               & \multicolumn{1}{c}{greet}              & \multicolumn{1}{c}{direct}             & \multicolumn{1}{c}{discuss}            & \multicolumn{1}{c}{walk}               & \multicolumn{1}{c}{eat}                & \multicolumn{1}{c}{phone}              & \multicolumn{1}{c}{purchase}           & \multicolumn{1}{c}{sit}                & \multicolumn{1}{c}{\begin{tabular}[c]{@{}c@{}}sit\\ down\end{tabular}} & \multicolumn{1}{c}{smoke}              & \multicolumn{1}{c}{\begin{tabular}[c]{@{}c@{}}take\\ photo\end{tabular}} & \multicolumn{1}{c}{\begin{tabular}[c]{@{}c@{}}walk\\ dog\end{tabular}} & \multicolumn{1}{c}{\begin{tabular}[c]{@{}c@{}}walk\\ together\end{tabular}} \\ \midrule
      \multicolumn{18}{c}{\textbf{\emph{fully supervised}}}                                                                                                                                                                     \\
                                                            & Newell~\etal~\cite{newell2016stacked}     & \cellcolor[HTML]{FFFFFF}19.52          & \cellcolor[HTML]{FFFFFF}15.53          & \cellcolor[HTML]{FFFFFF}13.88          & \cellcolor[HTML]{FFFFFF}17.14          & \cellcolor[HTML]{FFFFFF}15.81          & \cellcolor[HTML]{FFFFFF}19.55          & \cellcolor[HTML]{FFFFFF}13.74 & \cellcolor[HTML]{FFFFFF}15.33          & \cellcolor[HTML]{FFFFFF}18.81          & \cellcolor[HTML]{FFFFFF}19.88          & \cellcolor[HTML]{FFFFFF}25.85          & \cellcolor[HTML]{FFFFFF}39.07                                          & \cellcolor[HTML]{FFFFFF}19.40          & \cellcolor[HTML]{FFFFFF}22.24                                            & \cellcolor[HTML]{FFFFFF}21.58                                          & \cellcolor[HTML]{FFFFFF}14.96                                               \\
      \midrule
      \multicolumn{18}{c}{\textbf{\emph{self-supervised + supervised post-processing}}}\\
                                                            & Jakab~\etal~\cite{jakabunsupervised} & \cellcolor[HTML]{FFFFFF}19.12          & \cellcolor[HTML]{FFFFFF}16.63          & \cellcolor[HTML]{FFFFFF}15.01          & \cellcolor[HTML]{FFFFFF}16.68          & \cellcolor[HTML]{FFFFFF}14.73          & \cellcolor[HTML]{FFFFFF}15.69          & \cellcolor[HTML]{FFFFFF}17.74          & \cellcolor[HTML]{FFFFFF}16.53          & \cellcolor[HTML]{FFFFFF}23.27          & \cellcolor[HTML]{FFFFFF}17.35          & \cellcolor[HTML]{FFFFFF}24.66          & \cellcolor[HTML]{FFFFFF}33.14          & \cellcolor[HTML]{FFFFFF}20.31          & \cellcolor[HTML]{FFFFFF}20.96          & \cellcolor[HTML]{FFFFFF}\textbf{17.77}         & \cellcolor[HTML]{FFFFFF}16.31                                               \\
      \midrule
      \multicolumn{18}{c}{\textbf{\emph{self-supervised (no post-processing)}}}\\

      \multirow{-2}{*}{}                                    & Ours \textit{3DHP prior}                                & \cellcolor[HTML]{FFFFFF}18.94          & \cellcolor[HTML]{FFFFFF}15.33          & \cellcolor[HTML]{FFFFFF}14.37          & \cellcolor[HTML]{FFFFFF}16.08          & \cellcolor[HTML]{FFFFFF}15.90          & \cellcolor[HTML]{FFFFFF}17.24          & \cellcolor[HTML]{FFFFFF}14.51          & \cellcolor[HTML]{FFFFFF}17.30          & \cellcolor[HTML]{FFFFFF}19.66          & \cellcolor[HTML]{FFFFFF}17.39          & \cellcolor[HTML]{FFFFFF}22.79          & \cellcolor[HTML]{FFFFFF}30.84          & \cellcolor[HTML]{FFFFFF}18.50          & \cellcolor[HTML]{FFFFFF}24.21          & \cellcolor[HTML]{FFFFFF}23.77          & \cellcolor[HTML]{FFFFFF}16.16                                  \\
      \cmidrule(lr){2-2}   
      \multirow{-2}{*}{}                                    & Ours \textit{H3.6M prior}                               & \cellcolor[HTML]{FFFFFF}\textbf{14.46} & \cellcolor[HTML]{FFFFFF}\textbf{11.40}          & \cellcolor[HTML]{FFFFFF}\textbf{10.39}          & \cellcolor[HTML]{FFFFFF}\textbf{11.85}          & \cellcolor[HTML]{FFFFFF}\textbf{11.26}          & \cellcolor[HTML]{FFFFFF}\textbf{13.72}          & \cellcolor[HTML]{FFFFFF}\textbf{11.85}          & \cellcolor[HTML]{FFFFFF}\textbf{12.02}          & \cellcolor[HTML]{FFFFFF}\textbf{14.42}          & \cellcolor[HTML]{FFFFFF}\textbf{12.90}          & \cellcolor[HTML]{FFFFFF}\textbf{17.01}          & \cellcolor[HTML]{FFFFFF}\textbf{25.71}          & \cellcolor[HTML]{FFFFFF}\textbf{14.35}          & \cellcolor[HTML]{FFFFFF}\textbf{18.67}          & \cellcolor[HTML]{FFFFFF}19.42          & \cellcolor[HTML]{FFFFFF}\textbf{11.90}                               \\
      \bottomrule
      \end{tabular}
}

\caption{{\bf Human landmark detection (full H3.6M).} Comparison on Human3.6M test set with a supervised baseline Newell~\etal~\cite{newell2016stacked}, and a self-supervised method~\cite{jakabunsupervised}. We report the MSE in pixels~\cite{h36m_pami} for each activity. We highlight the minimum error across all models in bold.}
\label{atab:pose-sota}
\end{table}

\clearpage
\section{Appearance and geometry factorization}\label{as:factor}

\clearpage
\section{Facial landmarks detections}\label{as:face_test}
\begin{figure*}[ht]
\centering
\input{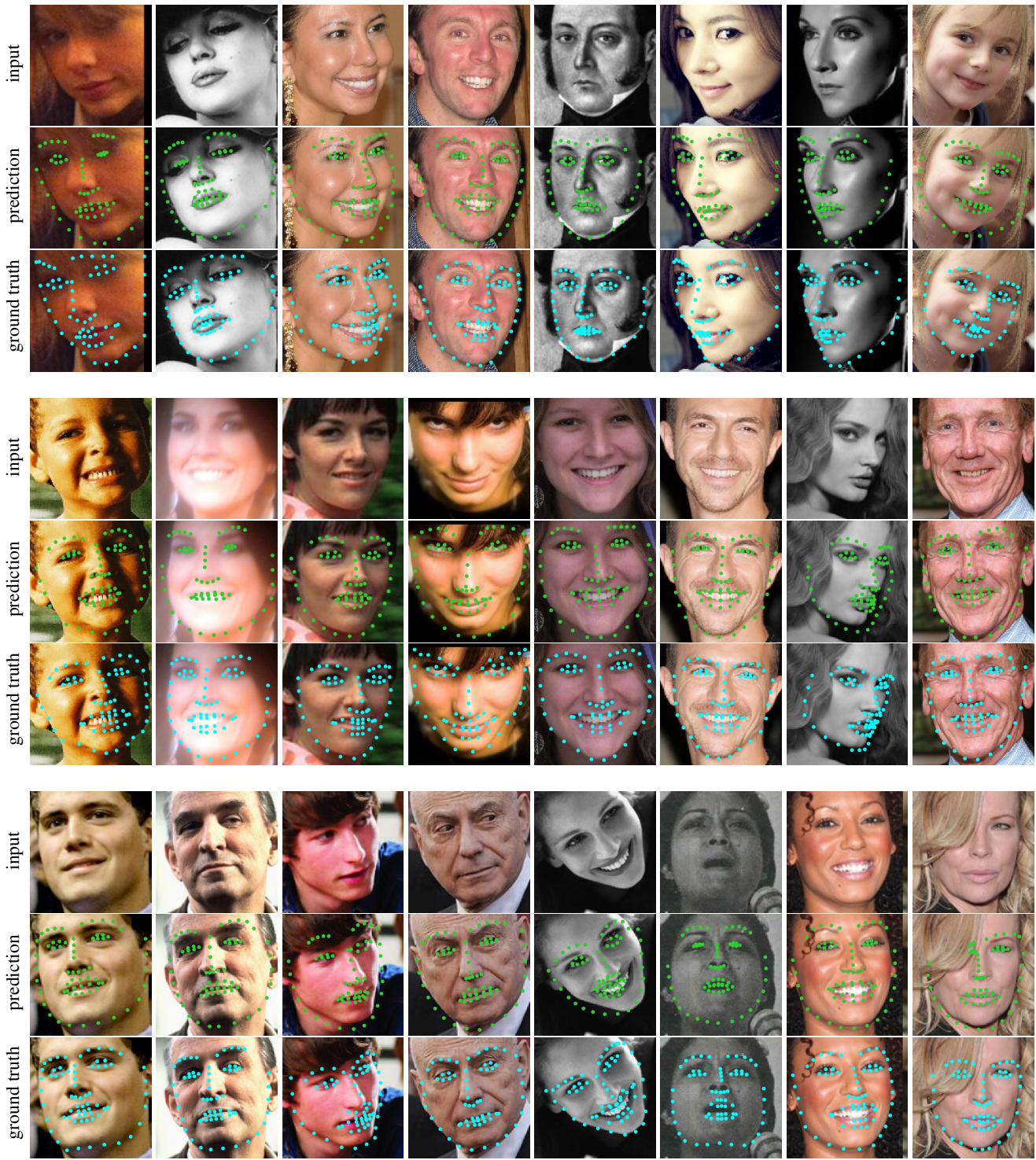}
\caption{\textbf{Facial landmark detections on 300-W.}
Randomly sampled predictions from 300-W test set. 
The model was trained with unlabelled images from VoxCeleb2 face videos dataset and unpaired landmarks sampled from MultiPIE dataset, hence shows significant generalization.
{\color{green} Green} markers denote our detections, {\color{cyan} blue} correspond to the ground truth.}
\label{af:}
\end{figure*}

\clearpage
\section{Human pose estimation}\label{as:human_test}
\subsection{Pose detection on Human3.6M}\label{as:}

\begin{figure*}[ht]
\centering

\input{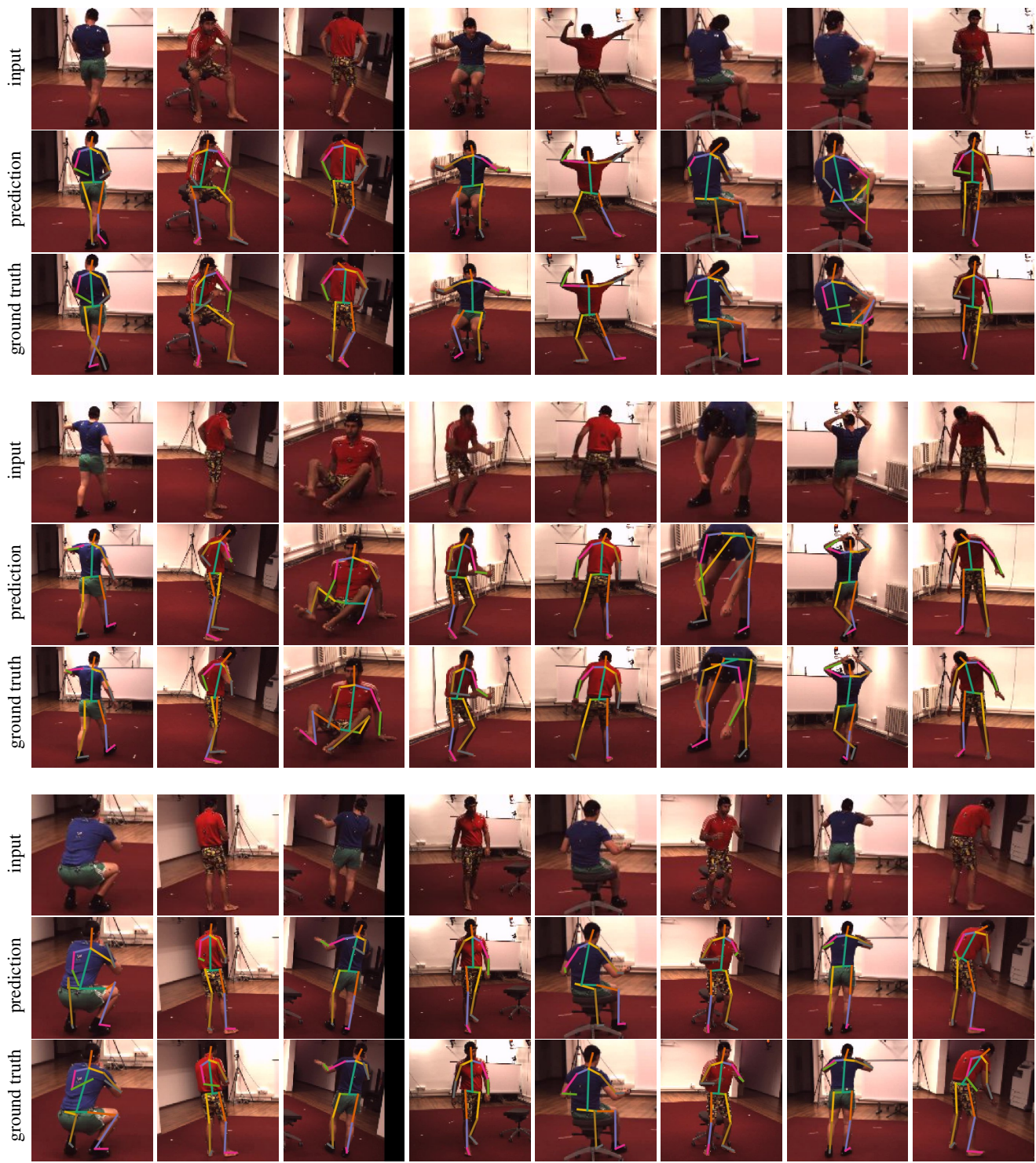}

\caption{\textbf{Pose estimation on Human3.6M.}
Randomly sampled results from Human3.6M test set.
The model is trained with unpaired images and skeletons from Human3.6M.
}
\label{f:}
\end{figure*}

\clearpage
\subsection{Pose detection on Simplified Human3.6M}\label{as:simple_human_test}

\begin{figure*}[ht]
\centering

\input{texgen/simple_human36m}

\caption{\textbf{Pose estimation on the Simplified Human3.6M.}
Randomly sampled results from the Simplified Human3.6M test set.
The model is trained with unpaired images and skeletons from Simplified Human3.6M.}
\label{f:}
\end{figure*}

\clearpage
\section{Landmarks detection on Cat Heads}\label{as:cat_test}

\begin{figure*}[ht]
\centering
\input{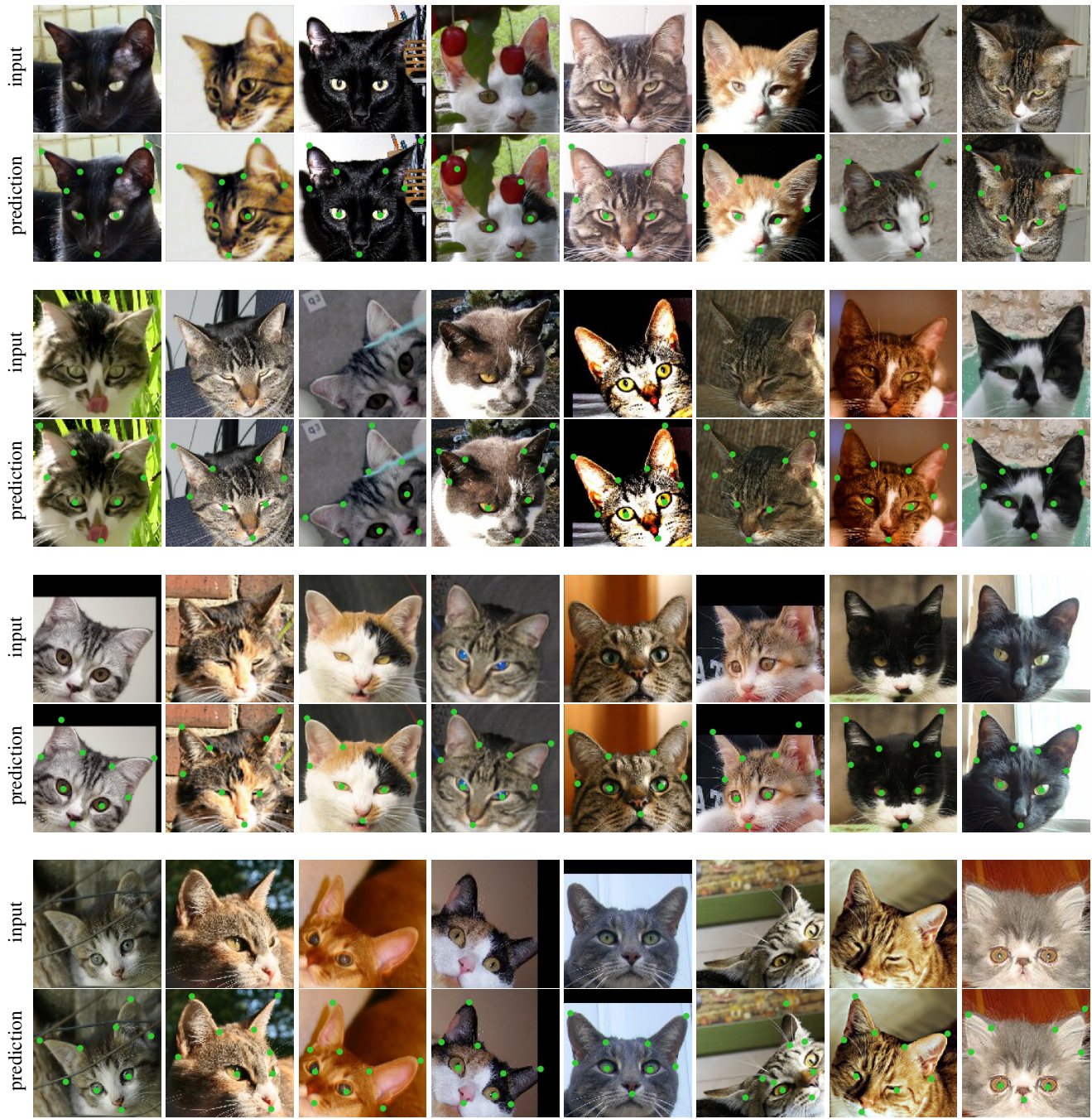}
\caption{\textbf{Landmark detections on Cat Head.}
Randomly sampled predictions on Cat Head test set.
}
\label{f:}
\end{figure*}

\clearpage
\section{Implementation details}\label{as:architectures}

\subsection{Training details}
The auto-encoder functions $\Phi$  and $\Psi$ and the discriminator $D$ are trained by optimizing the overall objective in eq. (5) of the main paper while setting $\lambda=10$ ($\eta$ is pre-trained using unpaired landmarks as detailed below).
We use the Adam optimiser \citep{kingma2014adam} with a learning rate of $2\cdot10^{-4}$, $\beta_1=0.5$ and $\beta_2=0.999$.
The batch size is set to 16 and the norm of the gradients is clipped to 1.0 for stability.

\subsection{Pre-training the function $\eta$}\label{as:pretrain}

The network $\eta$ mapping the skeleton image $\by$ to its corresponding keypoint locations $\bp$ is pre-trained before optimizing the overall objective (eq. (5) of the main paper).
This is done by using the unpaired pose samples $\{ \bar \bp_j \}_{j=1}^M$ and by optimizing the loss
$
\frac{1}{M}\sum_{j=1}^M \mathcal{L}(\eta|\bar \bp_j)
$
where
\begin{equation}\label{e:loss_reg}
\mathcal{L}(\eta|\bar \bp)
=
\| \eta \circ \beta(\bar\bp) - \bar\bp \| ^2
\end{equation}
is a simple $\ell^2$ regression loss.

During the optimization of the overall objective, the function $\eta$ is further fine-tuned by minimizing the same loss plus~\cref{e:loss_reg} an additional term
$
   \mathcal{L}(\eta| \by)
   =
   \lambda' \| \beta \circ \eta(\by) - \by \| ^2,
$
where $\by$ is a reconstructed pose (see fig. 2 of the main paper).
The latter ensures that network $\eta$ works for poses that appear in the videos but not necessarily in the pose prior.
The two terms are balanced by the coefficient $\lambda'$.
After fine-tuning $\eta$, we noticed that it loses some of its ability to distinguish between frontal and dorsal views of human body (which is fairly ambiguous given only a skeleton image as input).
We correct its predictions by using the pre-trained version of $\eta$ at \cref{e:loss_reg} to determine the orientation of human body.

The function $\eta$ is designed as a neural network that converts the skeleton image $\by$ into $K$ heatmaps.
The locations of keypoints are further obtained as in~\cite{jakabunsupervised} by converting each heatmap into a 2D probability distribution.
The expectation of this probability distribution corresponds to the location of the keypoints.
The spatial coordinates are normalised to the $[-1,1]$ range and we set $\gamma = \frac{1}{0.04}$ in eq. (2) of the main paper.
The function is learned by minimizing the loss introduced above with $\lambda'=0.1$.

\subsection{Note on a second cycle constraint and discriminator}
Standard CycleGAN~\cite{zhu2017unpaired} enforces two cycle constraints $\Psi \circ \Phi(\bx) \approx \bx$ and $\Phi \circ \Psi (\by) \approx \by$.
Our model implements a conditional version of the first, while the second can be written as $\Phi(\Psi(\bar\by,\bx')) \approx \bar\by$.
CycleGAN also utilizes a discriminator $D_\mathcal{X}$ on images $\hat \Psi(\by)$ generated from skeletons to match their distribution to images $\bx$; the same discriminator applies here, except that images are generated conditionally $\Psi(\bar\by,\bx')$ and they are tested against the distribution of images $\bx$ from the same video, so $\mathcal{D}_\mathcal{X}(\Psi(\bar\by,\bx'),\bx')$ is conditional too.
Our ablation study shows that the additional cycle constraint and discriminator leads to worse performance, so we do not include them in our final version of the model.

\subsection{Architectures}
\begin{figure*}[ht]
\centering
\includegraphics[width=0.57\textwidth]{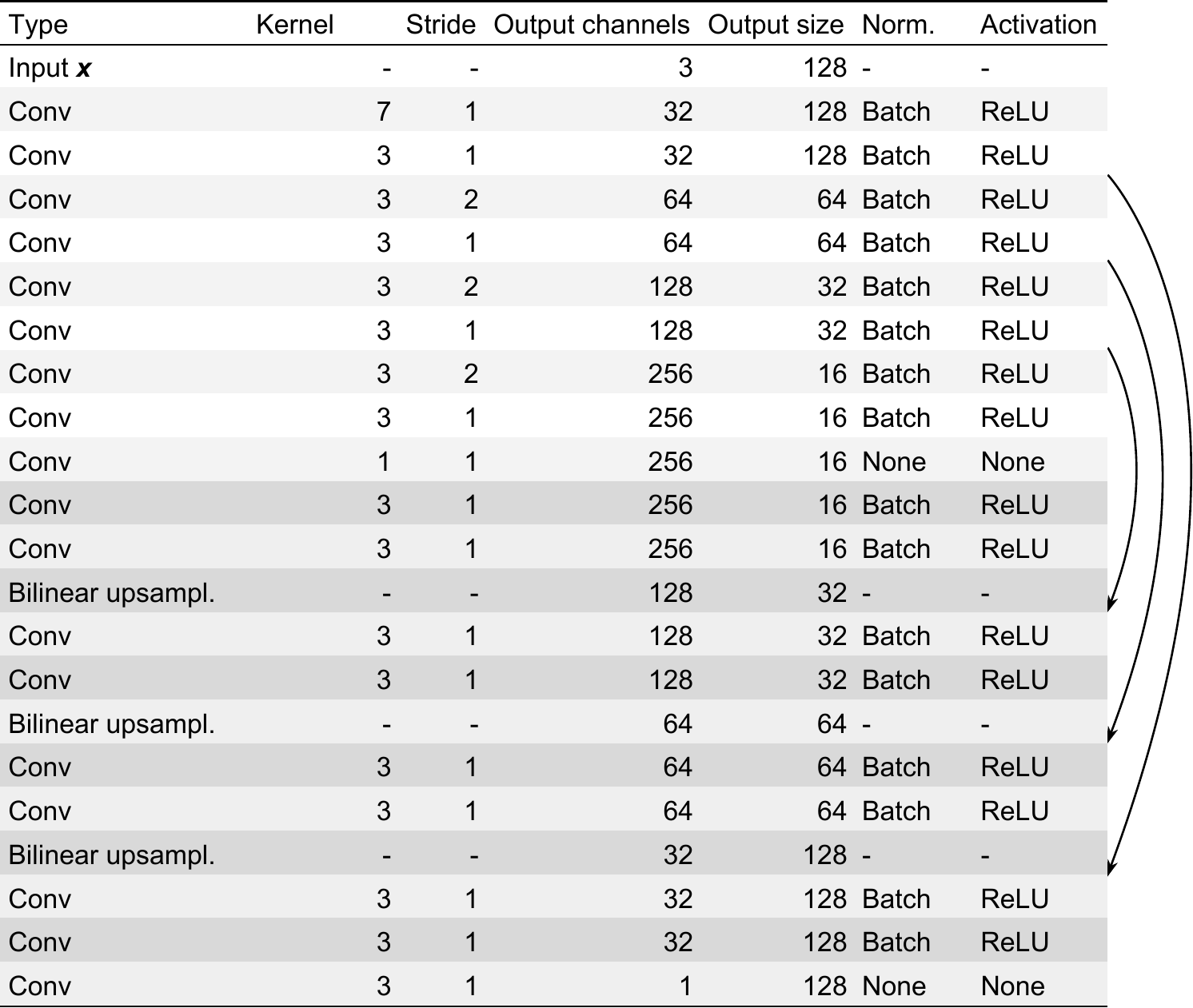}
\caption{\textbf{Image encoder $\Phi$.}
The network is based of the encoder and decoder network from~\cite{jakabunsupervised}.
Arrows on the side denote skip connections that are concatenated to the other input.
}
\label{f:im_enc}
\end{figure*}

\begin{figure*}[ht]
\centering
\includegraphics[width=0.8\textwidth]{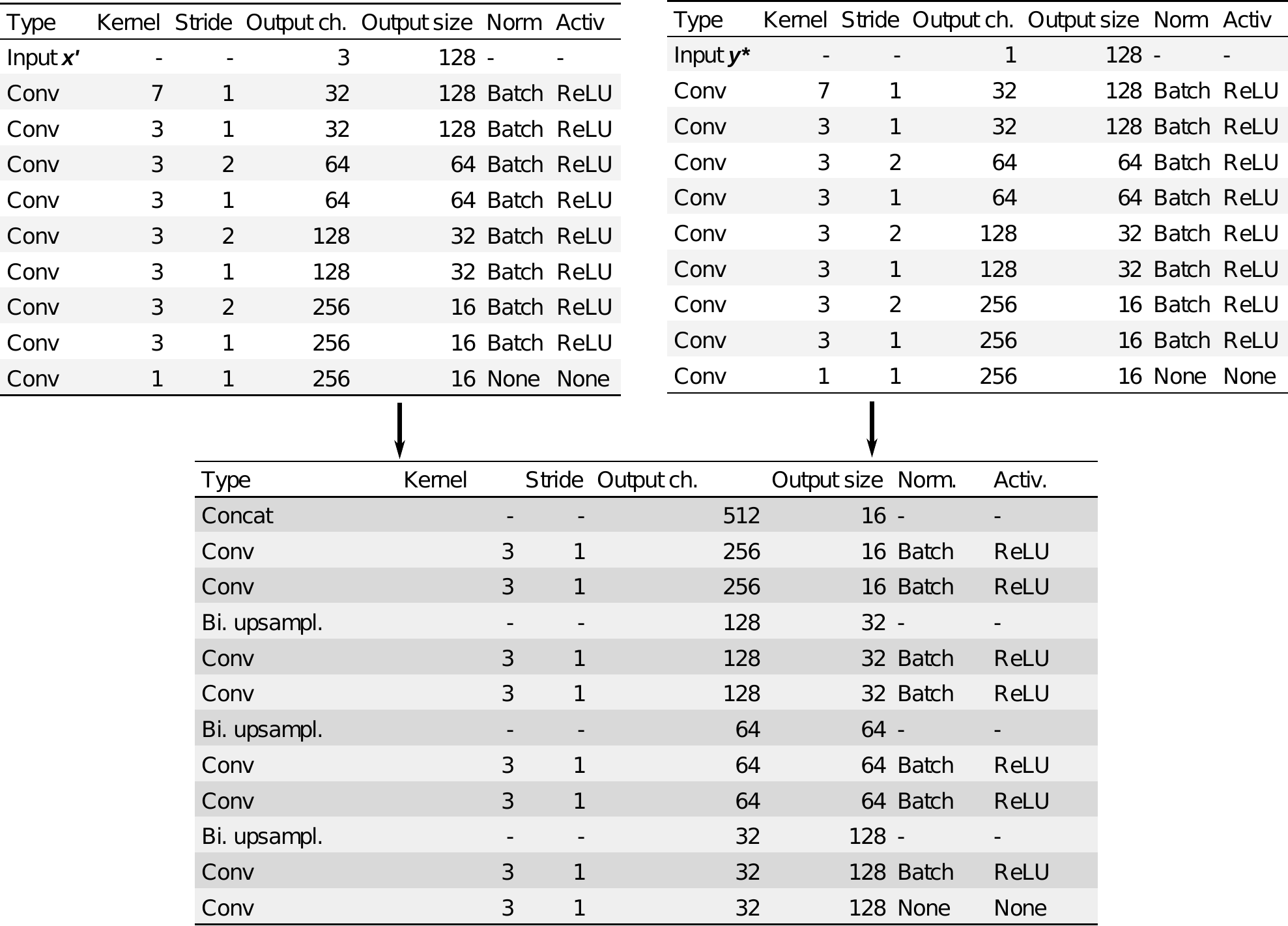}
\caption{\textbf{Image decoder $\Psi$.}
Image encoder first processes the conditioning image $\bx'$ and the skeleton $\by^*$ in two separate independent branches before it concatenates them into a single stream. The design follows~\cite{jakabunsupervised}.
}
\label{f:im_dec}
\end{figure*}

\begin{figure*}
\centering
\includegraphics[width=0.57\textwidth]{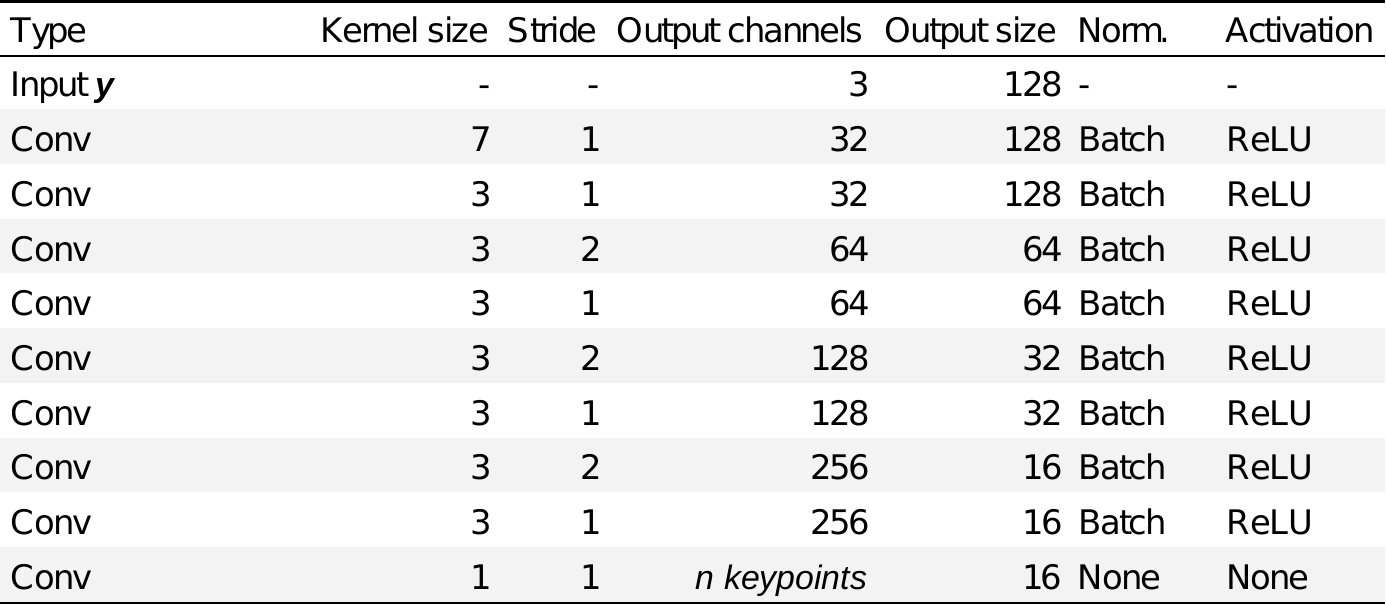}
\caption{\textbf{Skeleton encoder $\eta$.} The architecture is based on the encoder from~\cite{jakabunsupervised}. The last layer has as many output channels as the number of keypoints to predict.}
\label{f:skel_enc}
\end{figure*}

\begin{figure*}[ht]
\centering
\includegraphics[width=0.57\textwidth]{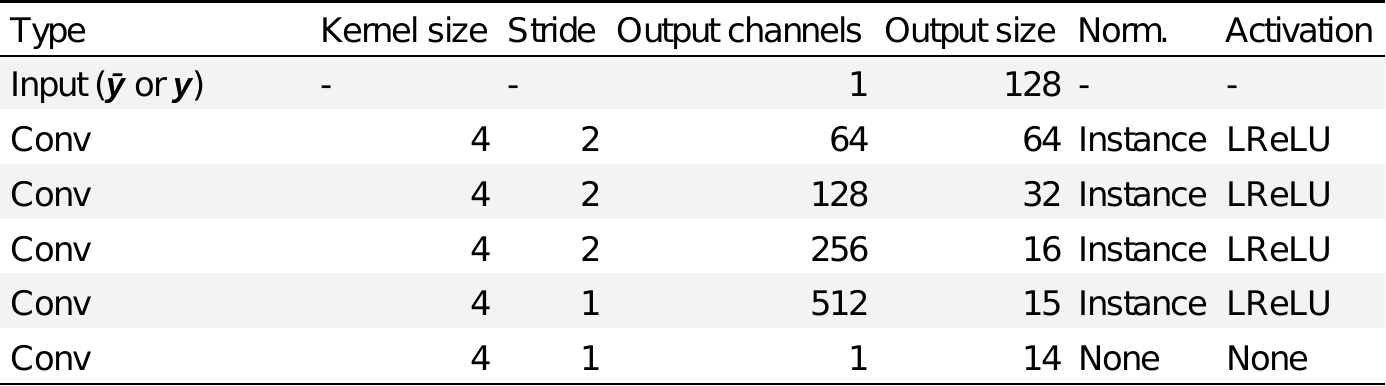}
\caption{\textbf{Skeleton discriminator $D_\mathcal{Y}$.}
The architecture follows~\cite{zhu2017unpaired}.
LReLU stands for Leaky Rectified Linear Unit~\cite{maas2013rectifier} that is used with $0.2$ negative slope.
Instance normalization~\cite{ulyanov2016instance} is used before every activation.
We use three such discriminators each for a different scale of the input image. We resize the input images by $1$, $\frac{1}{2}$, and $\frac{1}{4}$ factors.
}
\label{f:skel_disc}
\end{figure*}

\begin{figure*}[ht]
\centering
\includegraphics[width=\textwidth]{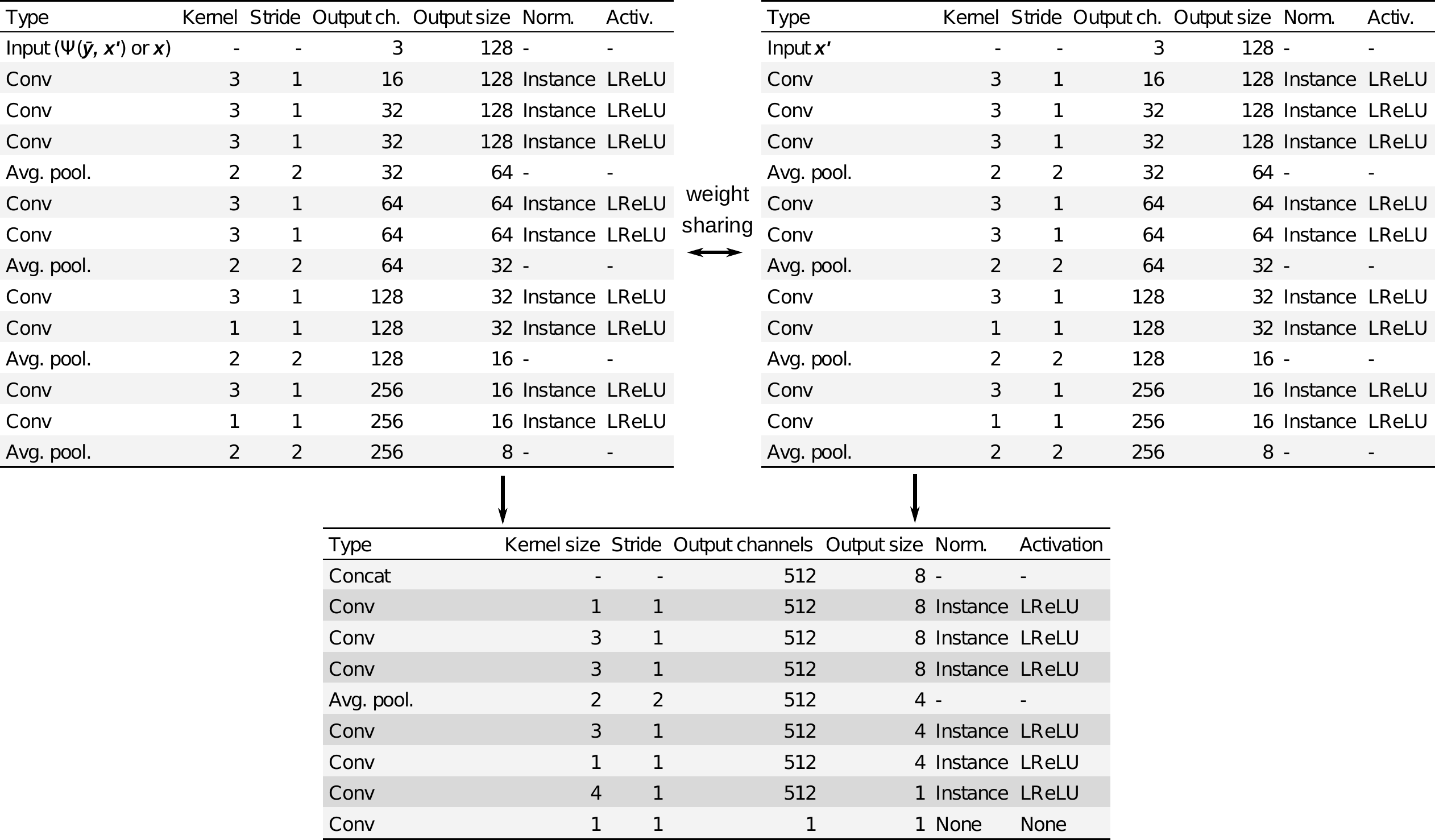}
\caption{\textbf{Conditional image discriminator $D_\mathcal{X}$.}
Conditional image discriminator starts with a Siamese architecture until the two streams are concatenated.
When the version without conditioning is required, the second branch in the Siamese part is simply omitted.
LReLU stands for Leaky Rectified Linear Unit~\cite{maas2013rectifier}. We set the negative slope to $0.2$. Every activation is preceded by instance normalization~\cite{ulyanov2016instance}. The architecture is loosely based on~\cite{karras2017progressive}.}
\label{f:im_disc}
\end{figure*}

\Cref{f:im_enc,f:im_dec,f:skel_enc,f:skel_disc,f:im_disc} provide detailed descriptions of network architectures used in experiments.

\clearpage

\end{document}